\documentclass[journal,lettersize]{IEEEtran}

\usepackage{color,xcolor}
\usepackage[sort&compress,numbers,comma]{natbib}

\newcommand{\minitab}[2][l]{\begin{tabular}{#1}#2\end{tabular}}
\usepackage{amsmath,amssymb,balance,bm,hyperref,multirow}
\hypersetup{hidelinks}
\usepackage{subfigure}
\usepackage{makecell}

\newcommand\eg{\emph{e.g.}}
\newcommand\ie{\emph{i.e.}}
\newcommand\etal{\emph{et al. }}

\ifCLASSINFOpdf
  \usepackage[pdftex]{graphicx}
  \graphicspath{{./images/}}
  \DeclareGraphicsExtensions{.pdf,.jpeg,.png}
\else
  \usepackage[dvips]{graphicx}
  \graphicspath{{./images/}}
  \DeclareGraphicsExtensions{.eps}
\fi

\usepackage{url}

\usepackage{algorithm,algorithmic}

\begin{document}

\title{Mask Attack Detection Using Vascular-weighted Motion-robust rPPG Signals}

\author{Chenglin Yao, \IEEEmembership{Student Member, IEEE}, Jianfeng Ren, \IEEEmembership{Senior Member, IEEE},\\ Ruibin Bai, \IEEEmembership{Senior Member, IEEE}, Heshan Du, Jiang Liu, Xudong Jiang, \IEEEmembership{Fellow, IEEE}
\thanks{Chenglin Yao, Jianfeng Ren, Ruibin Bai, Heshan Du are affiliated with the School of Computer Science, University of Nottingham Ningbo China. (email: Jianfeng.Ren@nottingham.edu.cn).}
\thanks{Jiang Liu is with the School of Computer Science, University of Nottingham Ningbo China, and with the Research Institute of Trustworthy Autonomous Systems and Department of Computer Science and Engineering, Southern University of Science and Technology, Shenzhen 518055, China.}
\thanks{Xudong Jiang is affiliated with School of Electrical \& Electronics Engineering, Nanyang Technological University, Singapore 639798.}
\thanks{This work was supported in part by the National Natural Science Foundation of China under Grant 72071116 and the Ningbo Municipal Bureau of Science and Technology under Grants 2019B10026.} 
}

\markboth{Journal of \LaTeX\ Class Files,~Vol.~0, No.~0, April~2022}%
{Shell \MakeLowercase{\textit{\etal}}: Bare Demo of IEEEtran.cls for IEEE Journals}

\maketitle

\begin{abstract}
Detecting 3D mask attacks to a face recognition system is challenging. Although genuine faces and 3D face masks show significantly different remote photoplethysmography (rPPG) signals, rPPG-based face anti-spoofing methods often suffer from performance degradation due to unstable face alignment in the video sequence and weak rPPG signals. To enhance the rPPG signal in a motion-robust way, a landmark-anchored face stitching method is proposed to align the faces robustly and precisely at the pixel-wise level by using both SIFT keypoints and facial landmarks. To better encode the rPPG signal, a weighted spatial-temporal representation is proposed, which emphasizes the face regions with rich blood vessels. In addition, characteristics of rPPG signals in different color spaces are jointly utilized. To improve the generalization capability, a lightweight EfficientNet with a Gated Recurrent Unit (GRU) is designed to extract both spatial and temporal features from the rPPG spatial-temporal representation for classification. The proposed method is compared with the state-of-the-art methods on five benchmark datasets under both intra-dataset and cross-dataset evaluations. The proposed method shows a significant and consistent improvement in performance over other state-of-the-art rPPG-based methods for face spoofing detection.
\end{abstract}


\IEEEpeerreviewmaketitle

\section{Introduction}

\IEEEPARstart{F}{ace} recognition systems have been widely deployed for identity authentication, whereas spoofing attack is still one of the major vulnerabilities to system security. 
There are three main types of spoofing to a face recognition system: 1) print attack, spoofing using a printed face image; 2) video-replay attack, spoofing using a piece of short live video with facial expressions; and 3) 3D mask attack, spoofing using a super-real 3D mask with skin-like materials. The first two types of attacks have been studied for years and many successful techniques \cite{boulkenafet2015face,chen2020attention,shen2019facebagnet,sun2020face,wang2021unsupervised} have been developed, while detecting 3D mask attacks still remains challenging due to the skin-like surface and precise 3D structures of a 3D mask.

In literature, researchers have developed many algorithms to detect 3D mask attacks, which can be categorized as: texture-based methods \cite{erdogmus2014spoofing,kose2013shape,steiner2016reliable}, motion-based methods \cite{shao2017deep,shao2018joint,siddiqui2016face}, shape-based methods \cite{tang2016shape,wang2018face}, rPPG-based methods \cite{birla2022patron,birla2022sunrise,li2016generalized,liu20163d,liu2018remote,liu2020temporal,liu2021multi,liu2022learning,nowara2017ppgsecure,yao2021rppgbased,yu2021transrppg}, methods based on deep convolutional neural networks \cite{george2020biometric,liu2019deep,liu2020disentangling,yu2020nasfas,liu2022contrastive}, and methods based on other liveness cues such as thermal signatures \cite{bhattacharjee2017what} and gaze~\cite{alsufyani2018biometric}.
Among these methods, the rPPG signal, which models the tiny periodic changes on skin color due to the heartbeat, shows good discriminative power to detect 3D mask attacks, since 3D masks are made from resin, plaster or silicone, which cannot produce such rPPG signals.

Nevertheless, methods based on rPPG signals face two challenges. One is the alignment error of face sequences. Faces are usually aligned through facial landmarks \cite{nowara2017ppgsecure,yao2021rppgbased,yu2021transrppg,niu2019robust,niu2020rhythmnet} to enhance the quality of the rPPG signals, whereas face landmarks are often difficult to locate precisely down to the pixel-wise level. For example, it is hard to precisely and consistently label the location of a nose tip, but the nose tip is often used in face alignment. Such a misalignment of facial landmarks can be tolerated to a certain extent in face recognition systems, but it greatly distorts the rPPG signal because the temporal rPPG signal from video frames is sensitive to their spatial positions. The other challenge is that the rPPG signal is weak and noisy. While the color change of rPPG signals is often out of the sensitivity of the human vision system, facial micro-motions and illumination variation significantly change the pixel values by a large amount, which results in low signal to noise ratio. These two challenges enormously affect the precision of rPPG signals. 
To address these challenges, three techniques are proposed in this paper. 

Firstly, to reduce the alignment error, a face-stitching algorithm is designed to align the faces in a video sequence to extract motion-robust rPPG signals, making use of not only the facial landmarks, but also facial keypoints. The proposed algorithm detects the facial keypoints down to the pixel-wise level by using the SIFT descriptor \cite{lowe2004distinctive}, which addresses the problem of ambiguous localization of facial landmarks in traditional facial alignment algorithms. At the same time, the facial landmarks are used as anchor points in video frames to avoid the problem of error propagation and address the drawback of unstable keypoint matching for two faces with a large pose difference. As the face motion between two consequent video frames in our application is small, an algorithm is proposed to accurately match keypoints, which minimizes not only the differences between feature representations, but also the spatial distance between two matched keypoints. 

Secondly, to enhance the rPPG signal and focus on the facial regions with rich blood vessels, a signal weighting mechanism based on the vascular density is proposed. Traditionally, rPPG signals are weighted either empirically \cite{yao2021rppgbased} or in a data-driven way by minimizing the evaluation error \cite{niu2020rhythmnet}, which lacks biological support. rPPG signals are originated from the periodical change of the skin color because of heartbeat \cite{wang2017algorithmic}, and hence it is conjectured that a facial region with rich blood vessels will result in a strong rPPG signal. Thus, larger weights should be given to regions with richer blood vessels when fusing the rPPG signals from different regions. The density of blood vessels is estimated by projecting the blood vessels of a biological specimen \cite{sun2018colour} to the face after proper alignments. 

Lastly, to explore the discriminant features in different color spaces, a feature fusion scheme is proposed to combine the rPPG signals in different color spaces. rPPG signals model the tiny color changes of skin pixels, and these tiny changes may exhibit different characteristics in different color channels of different color spaces, which could form robust patterns for the face spoofing detection. Thus, the rPPG signals from different color spaces are combined to form a spatial-temporal feature representation to enhance the rPPG signal and boost the classification performance.

In the proposed framework, the pixel-wisely aligned faces are divided into regions of interest (ROIs), and the enhanced rPPG signals are extracted from each ROI, and encoded as a spatial-temporal representation. To improve the generalization capability of the classifier, an EfficientNet with a Gated Recurrent Unit (GRU) is designed to extract reliable spatial-temporal features for classification, where the lightweight EfficientNet Blocks are designed for spatial feature learning and the GRU is designed for temporal feature learning. The proposed Vascular-weighted
Motion-robust rPPG (VMrPPG) is validated on five publicly available datasets for detecting spoofing attacks, and demonstrates a superior performance compared with the state-of-the-art rPPG-based algorithms for face spoofing detection.

Our contributions can be summarized as follows: 1) An image stitching algorithm on both facial landmarks and facial keypoints is proposed to reduce the alignment error so as to improve the robustness of face alignment, which makes use of the temporal consistency of keypoints to align the face reliably down to the pixel-wise level. 2) Inspired by the characteristics of rPPG signals, a weighted spatial-temporal representation based on the distribution of blood vessels is proposed to highlight the face regions with rich blood vessels. 3) To make full use of the tiny color changes of skin pixels, a color-fusion scheme is proposed to combine the rPPG signals in different color spaces. 4) Lastly, to improve the generalization capability, a lightweight EfficientNet with GRU is proposed to detect the spoofing face.

\section{Related Work}
In literature, many face anti-spoofing methods employ liveness cues such as textures, motions, shapes, and rPPG signals. In this section, the spoofing detection methods based on the appearances of 3D mask attacks will be reviewed first, followed by the methods for rPPG signal enhancement.

\subsection{3D Mask Attack Detection}
\noindent\textbf{Texture-based Methods} exploit the difference in texture pattern between spoofing faces and genuine faces for spoofing detection \cite{jia2020survey}. Texture descriptors such as Local Binary Pattern (LBP) \cite{kose2013shape,erdogmus20133dmad,erdogmus2014spoofing} and Binarized Statistical Image Features (BSIF) \cite{kannala2012bsif} have been used for face anti-spoofing. Kose \etal \cite{kose2013shape} introduced a multi-scale LBP on both RGB images and depth images to detect the abnormality of masked faces. Erdogmus and Marcel extended it to other descriptors such as modified LBP, transitional LBP, and direction-coded LBP \cite{erdogmus20133dmad,erdogmus2014spoofing}. LBP serves as a baseline method for many datasets, \eg, 3DMAD \cite{erdogmus20133dmad} and HKBU-Mars \cite{liu2016hkbu-mars-v2,liu2018remote}. 
These handcrafted features can recognize unique textures on facial masks, but they often have limited discriminant power when faced with different illumination conditions or mask materials. 

\noindent\textbf{Motion-based Methods} mainly focus on unconscious subtle facial expressions such as eye blink, mouth movement, and facial muscle contraction, which cannot be observed on some rigid facial masks. 
Siddiqui \etal \cite{siddiqui2016face} encoded textures using LBP features at multiple scales and extracted micro-movement patterns in consecutive frames via Histogram of Oriented Optical Flows. Shao \etal \cite{shao2017deep} extracted the subtle facial motion using a VGG Net. Liu \etal~\cite{liu2018learning} introduced the CNN-RNN architecture for facial motion feature extraction and classification. 

\noindent\textbf{Shape-based Methods} employ the differences in 3D face structures to detect spoofing attacks. Tang and Chen \cite{tang2016shape} applied Principal Curvature Measures and meshed SIFT-based features to the face spoofing detection. Hamdan and Mokhtar \cite{hamdan2018self} combined features from Legendre Moments Invariants and Linear Discriminant Analysis as the liveness cue, and classified them using Maximum Likelihood Estimation. These methods apply image transformation to extract shape features, while Wang \etal \cite{wang2018face} obtained geometry features by reconstructing a 3D morphable model from RGB images, and combined them with LBP features under both handcrafted fusion and VGG-generated fusion.

\noindent\textbf{rPPG-based Methods} extract the liveness clues of heartbeats through RGB cameras using remote photoplethysmography (rPPG) technology. rPPG signals are initially extracted for heart rate estimation \cite{li2014remote} and before long employed for face anti-spoofing. Under natural environments, researchers have noticed that the major challenge is the background noise. To tackle this problem, Liu \etal \cite{liu20163d} divided the whole face into ROIs, and constructed the local rPPG correlation model, in which the phase and period information of rPPG signals from different ROIs are utilized as the liveness cues. This work is improved by using rPPG correspondence features \cite{liu2018remote} and multi-channel correspondence features \cite{liu2021multi} to differentiate genuine faces from spoofing attacks. \cite{liu2020temporal,liu2022learning}, the signal similarity between neighboring ROIs is further extended to three features, amplitude, gradient, and phase, measured by Euclidean distance, normalized cross correlation (NCC) metric, and dot production. In PATRON \cite{birla2022patron}, the respiratory signal is extracted from the rPPG signal as an auxiliary liveness cue. Another rPPG feature descriptor for face anti-spoofing is the long-term statistical spectral (LTSS) \cite{heusch2018pulse} which employs the first and second order statistics of the frequency spectrum of a signal. Its multi-scaled version (MS-LTSS) is fit to a Contextual Patch-based Convolutional Neural Network (CP-CNN) \cite{lin2019face} to obtain a better performance. The CNN-RNN classifier \cite{liu2018learning}, C(2+1) Network \cite{liu2022learning} and the transformer architecture \cite{yu2021transrppg} have also been adopted to extract the temporal information embedded in rPPG signals. 

\noindent\textbf{CNN-based Methods} extract liveness clues directly from raw face sequences \cite{yu2023deep}. Liu \etal \cite{liu2019deep} introduced a Deep Tree Network to cluster the video frames into sub-groups via the Tree Routing Unit and classify them using the Supervised Feature Learning module. George \etal \cite{george2020biometric} designed a Multi-channel Convolutional Neural Network to fuse the clues from multiple synchronized cameras using Domain Specific Units. To address the problem of insufficient samples for spoof traces, the Generative Adversarial Network (GAN) training strategy has been adopted recently \cite{liu2020disentangling}. Liu \etal introduced a Spoof Trace Disentanglement Network (STDN), in which the spoof traces are encoded into a hierarchical representation by the generator \cite{liu2020disentangling}. The extended work, STDN+ \cite{liu2023spoof}, explicitly estimates the spoofing-related patterns on the trace modeling. For multi-modal inputs, Liu \etal \cite{liu2021face} designed a Modality Translation Network (MT-Net) for the generator to map the patterns from different modalities and a Modality Assistance Network (MA-Net) for feature translation between different modalities. Qin \etal \cite{qin2022meta} designed a meta-teacher optimization framework to supervise the process of learning rich spoofing cues. To adapt the face spoofing detection to different scenarios in an automatic way, Yu \etal \cite{yu2020nasfas} developed a Neural Architecture Search (NAS) for face anti-spoofing, named NAS-FAS. Although the searched neural networks cannot outperform the state-of-the-art expert-designed networks, its high-level abstraction is promising.

\noindent\textbf{Methods Based on Other Liveness Cues} have also been developed. Agarwal \etal \cite{agarwal2017face} declared that the thermal imaging spectrum shows a predominant power to detect 3D mask attacks whereas such technology is costly. Directing gaze information to build behavior patterns has been shown effective to resist face spoofing attacks \cite{alsufyani2018biometric}. 

\subsection{Remote Photoplethysmography Signal}
An rPPG signal is a set of complex and weak signals of heartbeats with noise. The rPPG signals have been used in many applications such as remote heart rate monitoring~\cite{niu2020rhythmnet,niu2019robust} and face anti-spoofing \cite{liu20163d,liu2018remote,liu2020temporal,liu2021multi}. To suppress the noise in rPPG signals, methods have been developed for denoising and signal enhancements \cite{haan2013robust,wang2017algorithmic,yao2021rppgbased}. One way to filter the noise is to utilize the correlation information of different color channels. 
To filter the noise in general situations, CHROM \cite{haan2013robust} is designed by utilizing the knowledge of color model, which can be robust to non-white illuminations. Method 2SR \cite{wang2016novel} detects the rPPG signals by tracking hue changes of the skin. Wang \etal \cite{wang2017algorithmic} improved this work by incorporating data-driven discovery and physiological properties of skin reflections. Another way is to handle the rPPG signals in the frequency domain. The energy terms within the frequency range of normal heartbeat contain most heartbeat information while those out of the range mostly contains noise. Based on this idea, Lovisotto \etal \cite{lovisotto2020seeing} enhanced the signal using a lowpass filter at 4 Hz with the Beat Separation algorithm. Yao \etal \cite{yao2021rppgbased} applied a bandpass filter with the cut-off frequency at 0.8 Hz - 3.3 Hz.

\section{Proposed Face Alignment via Image Stitching}
\label{sec:joint-align}
\subsection{Motivations of Face Alignment via Image Stitching}

As discussed, existing rPPG-based spoofing detection methods face two major challenges, face alignment errors and weak rPPG signals. Face alignment errors are caused by ambiguity of facial landmarks and localization errors of landmarks. The ambiguity arises from the fact that facial landmarks may not be precisely and consistently annotated at a pixel-wise level in the first place. Face recognition systems can tolerate face alignment errors to a certain extent, but the errors significantly affect the quality of extracted rPPG signals. 
Different from landmarks, the keypoints in SIFT feature space~\cite{lowe2004distinctive} can be robustly detected in a pixel-wise precision, which partially addresses the problem of ambiguity annotation of facial landmarks. However, keypoint-based face alignment has its own challenges: 1) there are only few matched keypoints when aligned directly to the reference face over a remarkable pose difference; 2) the errors are inevitably propagated through repeating alignments between successive video frames.

To tackle these challenges, SIFT keypoints and facial landmarks are jointly utilized for precise face alignment.  
The SIFT keypoints are utilized for pixel-wise matching and the facial landmarks serve as anchor points to guide the matching, so that the error propagation can be minimized. The proposed joint face alignment for rPPG signal extraction consists of three phases. Firstly, to match keypoints between successive frames of a face video, a matching mechanism is developed to utilize both spatial similarity and feature similarity for the detected SIFT keypoints. Secondly, the affine transform is applied to align each face to the template. Finally, a landmark-anchored face stitching method is proposed, in which SIFT features are used to align a face indirectly to the template through a set of intermediate video frames, to address the challenges of few matched keypoints between two faces of a large pose difference. Landmarks are detected and matched directly in most frames, which serve as anchor points to stop the error prorogation in successive matching of keypoints. A dynamic programming method is developed to derive the set of intermediate frames with the minimal alignment error, through which a chain of affine transforms is derived to stably and precisely align the face to the template. 

\subsection{Keypoint Matching by Maximizing both Spatial and Feature Similarities}
For face alignment between two successive frames in a video sequence, two matched keypoints should be spatially close, since the head movements between two successive frames are usually insignificant. To make good use of spatial similarity and feature similarity, a keypoint matching mechanism is designed to take both into consideration. 
Given a face video sequence, a face detection tool SeetaFace \cite{wu2017funnel} is utilized firstly to crop the face region in each frame. Taking the first frame as the template, the subsequent frames are aligned to the template. For offline video sequences, the template can be the middle frame to mitigate the error propagation. The facial landmarks are also detected by SeetaFace and the keypoints are extracted by SIFT algorithm \cite{lowe2004distinctive}, with a feature vector $\bm{f} \in \mathbb{R}^{128}$ and a 2D space vector $\bm{p}=[x,y] \in \mathbb{N}^{2}$.

The keypoint matching for faces in two frames (namely, the reference face and the query face) consists of two steps: initial matching to restrict the search space and fine matching.
In \cite{lowe2004distinctive}, the k-Nearest-Neighbor (kNN) algorithm is applied on features alone to find the initial matching, while both spatial and feature distances are utilized in the proposed method. Denote the feature and spatial distances of two nearest neighbors in the query face
to each keypoint 
in the reference face as $d_{F1}$, $d_{S1}$ and $d_{F2}$, $d_{S2}$ after proper range normalization. The fused distance is formulated as their Euclidean distance:
\begin{equation}
\label{eqn:dmi}
	d_{Mi}=\sqrt{d_{Fi}+d_{Si}} \quad s.t. \quad i \in \{1,2\}.
\end{equation}
An initial match is granted when $d_{M1}\le \delta d_{M2}$. The threshold $\delta$ is adaptable to different camera configurations for a balance of keypoint quantity and matching speed.

In the second step, \ie, fine matching, the feature distance $d_{F}$ and spatial distance $d_{S}$ for each initial match are calculated. The initial feature and spatial distance sets of all initial matches, denoted as $\mathcal{D}_{F}$ and $\mathcal{D}_{S}$, are assumed to follow the Gaussian distribution and the two distributions are independent. The joint distribution can be modeled as:
\begin{align}
\label{eqn:joint-dis}
G(d_{F}, d_{S})&=G(d_{F})\cdot G^{\lambda}(d_{S}) \nonumber \\ 
 &\propto -\frac{(d_{F}-\mu_{F})^2}{2\sigma_{F}^2}-\lambda \frac{(d_{S}-\mu_{S})^2}{2\sigma_{S}^2},	
\end{align}
where $\sigma_{F}$, $\mu_{F}$ and $\sigma_{S}$, $\mu_{S}$ are the standard deviation and mean of $\mathcal{D}_{F}$ and $\mathcal{D}_{S}$, respectively. $\lambda$ is a weight to balance two distances. An acceptance rate $\alpha \in (0,1)$ is defined and the top $\alpha$ initial matches ordered by the joint distribution are the fine matches. 
The proposed method can accurately and efficiently derive the set of matched keypoints from face sequences.

\subsection{Face Alignment via Affine Transform}
\label{sec:affine}
The out-plane rotation between two successive frames in a face video is negligible. Thus, the mapping between two successive frames is modeled as an affine transform, following the same design as in \cite{geng2011face}. For each matched keypoint pair, the affine projection from the query face $\bm{v}=[x,y,1]^\mathsf{T}$ to the reference face $\bm{v'}=[x',y',1]^\mathsf{T}$ is formulated as:
\begin{equation}
	\begin{bmatrix}
		x' \\
		y' \\
		1
	\end{bmatrix}=\bm{P}
	\begin{bmatrix}
		x \\
		y \\
		1
	\end{bmatrix}=
	\begin{bmatrix}
		p_{11} & p_{12} & p_{13} \\
		p_{21} & p_{22} & p_{23} \\
		0 & 0 & 1
	\end{bmatrix}
	\begin{bmatrix}
		x \\
		y \\
		1
	\end{bmatrix},
	\label{eqn:affine-1}
\end{equation}
where the projection matrix $\bm{P}$ consists of $6$ coefficients to solve. When the number of matched keypoints is sufficient, the Least Mean Square method \cite{geng2011face} can be applied to find the projection matrix $\bm{P}$. 
When the amount of matched keypoints is not sufficient to compute a transformation matrix, the matched facial landmarks are utilized alternatively. 

\subsection{Landmark-anchored Face Stitching}
\subsubsection{Face Alignment Using Both Keypoints and Landmarks}
Facial landmarks such as nose tips are difficult to annotate consistently to the pixel-wise precision. Traditional face alignment through annotated landmarks hence may introduce errors and distort the weak and noise-sensitive rPPG signal. SIFT keypoints could be matched precisely to the pixel-wise level. However, SIFT keypoint matching for two faces over a large pose difference may fail and lead to very few matched keypoints. In this paper, we propose to align the query face to the reference face through a series of intermediate faces mainly using SIFT keypoints, with landmarks serving as anchor points to prevent the error propagation in excessive intermediate matches. As summarized in Alg. \ref{alg:dp-fa}, the proposed face alignment method searches for intermediate faces to minimize the alignment error by utilizing a dynamic programming method.

For all matched feature points between the reference face and the query face, the alignment error is defined as the Euclidean distance between the projected points from the query face and those of the reference face. Specifically, denote the reference face as $\mathcal{F}_{1}$ and the query face as $\mathcal{F}_{k}$, $k\geq2$, $k\in\mathbb{N^{+}}$. We first aligns the query face to an intermediate face $\mathcal{F}_{\hat{k}}$, $1\le\hat{k}<k$,  
and then indirectly to the reference face. 
As both SIFT keypoints and facial landmarks are utilized, the total alignment error $\mathcal{L}(1,k)$ is defined as the sum of the alignment error using SIFT keypoints $\mathcal{L}^{K}(1,k)$ and the alignment error using facial landmarks $\mathcal{L}^{L}(1,k)$:
\begin{equation}
	\mathcal{L}(1,k)=\mathcal{L}^{K}(1,k)+\mathcal{L}^{L}(1,k).
	\label{eqn:dp-general}
\end{equation}

These two types of errors are calculated differently. Firstly, the alignment error $\mathcal{L}^{K}(1,k)$ using keypoints is derived. When the pose difference between $\mathcal{F}_{k}$ and $\mathcal{F}_{1}$ is large, there may be too few matched keypoints to directly align $\mathcal{F}_{k}$ to $\mathcal{F}_{1}$. In this case, an intermediate face $\mathcal{F}_{\hat{k}}$ is used during the alignment between $\mathcal{F}_{k}$ and $\mathcal{F}_{1}$, where $\mathcal{F}_{1}$ is first aligned to $\mathcal{F}_{\hat{k}}$  and then $\mathcal{F}_{\hat{k}}$ is aligned to $\mathcal{F}_{k}$. Matched keypoints may be different between $(\mathcal{F}_{1},\mathcal{F}_{\hat{k}})$ and $(\mathcal{F}_{\hat{k}},\mathcal{F}_{k})$. In this case, the keypoint alignment error from $\mathcal{F}_{1}$ to $\mathcal{F}_{k}$ through an intermediate $\mathcal{F}_{\hat{k}}$ is estimated as follows:
\begin{equation}
	\mathcal{L}^{K}(1,k)=\mathcal{L}^{K}(1,\hat{k})+\sqrt{\frac{1}{m}\sum_{j=1}^{m}|\bm{P}_{\hat{k},k}\bm{v}_{k,j}^{K}-\bm{v}_{\hat{k},j}^{K}|^{2}},
	\label{eqn:kp-general}
\end{equation}
where the first term is the recursive definition of the alignment error from $\mathcal{F}_{1}$ to $\mathcal{F}_{\hat{k}}$ using keypoints, and the second term is the error of aligning $\mathcal{F}_{k}$ to $\mathcal{F}_{\hat{k}}$ using keypoints.
$\bm{v}_{k,j}^{K}$ represents the coordinate vector of the $j$-th matched SIFT keypoint in  $\mathcal{F}_{k}$, $\bm{v}_{\hat{k},j}^{K}$ is the corresponding keypoint in $\mathcal{F}_{\hat{k}}$ and $\bm{P}_{\hat{k},k}$ is the projection matrix from $\bm{v}_{k,j}^{K}$ of $\mathcal{F}_{k}$ to $\bm{v}_{\hat{k},j}^{K}$ of $\mathcal{F}_{\hat{k}}$, which can be estimated as outlined using Eqn. \eqref{eqn:affine-1} in Section \ref{sec:affine}.

Secondly, we derive the alignment error $\mathcal{L}^{L}(1,k)$ using landmarks. As the facial landmarks could be detected in almost all the frames, the alignment error through landmarks can be defined directly as follows:
\begin{equation}
	\mathcal{L}^{L}(1,k)=\sqrt{\frac{1}{n}\sum_{j=1}^{n}|\bm{P}_{1,k}\bm{v}^{L}_{k,j}-\bm{v}^{L}_{1,j}|^{2}},
	\label{eqn:ldmk-general}
\end{equation}
where $\bm{v}_{k,j}^{L}$ and $\bm{v}_{1,j}^{L}$ represent the coordinate vector of the $j$-th facial landmark out of $n$ from the query face $\mathcal{F}_{k}$ and the reference face $\mathcal{F}_{1}$, respectively. $\bm{P}_{1,k}$ is the projection matrix to align $\mathcal{F}_{1}$ and $\mathcal{F}_{k}$. When an intermediate face $\mathcal{F}_{\hat{k}}$ is used, 
the projection matrix $\bm{P}_{1,k}$ can be estimated as follows,
\begin{equation}
    \bm{P}_{1,k}=\bm{P}_{1,\hat{k}}\cdot\bm{P}_{\hat{k},k}, 
    \label{eqn:dp-projection}
\end{equation}
where $\bm{P}_{1,\hat{k}}$ and $\bm{P}_{\hat{k}, k}$ represent the projection matrices from $\mathcal{F}_{\hat{k}}$ to $\mathcal{F}_{1}$ and from $\mathcal{F}_{k}$ to $\mathcal{F}_{\hat{k}}$, respectively. 

Finally, by integrating Eqn.~(\ref{eqn:dp-general})-(\ref{eqn:dp-projection}), the alignment error from $\mathcal{F}_{1}$ to $\mathcal{F}_{k}$ through an intermediate face $\mathcal{F}_{\hat{k}}$ is calculated as,
\begin{equation}
\begin{aligned}
\mathcal{L}(1,k) &= \mathcal{L}^{K}(1,\hat{k})+ \sqrt{\frac{1}{m}\sum_{j=1}^{m}|\bm{P}_{\hat{k},k}\bm{v}_{k,j}^{K}-\bm{v}_{\hat{k},j}^{K}|^{2}} \\
&+ \sqrt{\frac{1}{n}\sum_{j=1}^{n}|\bm{P}_{1,\hat{k}}\bm{P}_{\hat{k},k}\bm{v}^{L}_{k,j}-\bm{v}^{L}_{1,j}|^{2}}.
\end{aligned}
\label{eqn:total_alignment_error}
\end{equation}
In the next subsection, a dynamic programming solution is proposed to find a set of intermediate faces to minimize the face alignment error $\mathcal{L}(1,k)$. 

\subsubsection{Dynamic Programming Solution for Face Alignment through Intermediate Faces}
The target here is to find a set of intermediate faces $\{\mathcal{F}_{k_{1}},\mathcal{F}_{k_{2}},...,\mathcal{F}_{k_{q}}\}$ between $\mathcal{F}_{1}$ and $\mathcal{F}_{k}$, so that $\mathcal{F}_{k}$ is aligned to $\mathcal{F}_{1}$ through this set of intermediate faces to minimize the error defined in Eqn.~(\ref{eqn:total_alignment_error}). An enumeration approach will result in $2^k$ different combinations of intermediate faces, but this method may have a lot of redundancy. Note that $\mathcal{L}^{K}(1,k')$ and $\bm{P}_{1,k'}$, $1<k'\leq k$, can be reused to reduce the complexity, which leads to the following dynamic programming solution summarized in Alg.~\ref{alg:dp-fa}.
\begin{algorithm} [!t]
	\caption{Dynamic programming for robust face alignment}
	\begin{algorithmic}[1]
		\renewcommand{\algorithmicrequire}{\textbf{Input:}}
		\renewcommand{\algorithmicensure}{\textbf{Output:}}
		\REQUIRE A set of keypoints $\{\bm{v}^{K}_{1},...,\bm{v}^{K}_{k}\}$ and a set of landmarks $\{\bm{v}^{L}_{1},...,\bm{v}^{L}_{k}\}$
		\ENSURE A set of projection matrices $\{\bm{P}_{1,2},...,\bm{P}_{1,k}\}$ 
		\FOR {$k'=2$ to $k$}
		\STATE Let the minimum total alignment loss $\mathcal{L}_{min} \leftarrow \infty$
		\STATE Let the index of intermediate face $\hat{k} \leftarrow 1$
		\FOR {$i=1$ to $k'-1$}
		\STATE Derive $\bm{P}_{i,k'}$ using $\bm{v}^{K}_{i},\bm{v}^{K}_{k'},\bm{v}^{L}_{i},\bm{v}^{L}_{k'}$ as outlined in Section \ref{sec:affine} 
		\STATE Update the alignment loss $\mathcal{L}(1,k')$ as defined in Eqn.~(\ref{eqn:total_alignment_error}) 
		\IF {$\mathcal{L}(1,k')<\mathcal{L}_{min}$}
		\STATE Update $\mathcal{L}_{min} \leftarrow \mathcal{L}(1,k')$
		\STATE Update $\hat{k}\leftarrow i$
		\ENDIF
		\ENDFOR
		\STATE Update $\mathcal{L}^K(1,k')$ using Eqn.~(\ref{eqn:kp-general}) with $\hat{k}$ 
		\STATE Update $\bm{P}_{1,k'} \leftarrow \bm{P}_{1,\hat{k}}\cdot\bm{P}_{\hat{k},k'}$ 
		\ENDFOR
		\RETURN $\{\bm{P}_{1,2},...,\bm{P}_{1,k}\}$
	\end{algorithmic}
	\label{alg:dp-fa}
\end{algorithm}

The key to this DP algorithm is to reuse $\mathcal{L}^{K}(1,k')$ and $\bm{P}_{1,k'}$, $1<k'\leq k$, during optimization. The projection matrices $\{\bm{P}_{1,2},\dots,\bm{P}_{1,k}\}$ and the minimal losses $\{\mathcal{L}^K(1,2),\dots,\mathcal{L}^K(1,k)\}$ are derived in sequence during each iteration of the outer loop. The time-complexity of this dynamic programming algorithm is $\mathcal{O}(k^2)$, which is much lower than $\mathcal{O}(2^k)$ for the native implementation using enumeration.

\begin{figure}[ht]
	\centering
	\subfigure[Landmark Only ]{\includegraphics[width=0.3\linewidth]{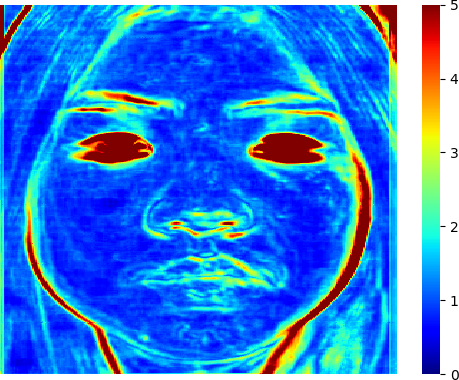}
		\label{fig:ldmk-vis}}
	\centering
	\subfigure[Keypoint Only ]{\includegraphics[width=0.3\linewidth]{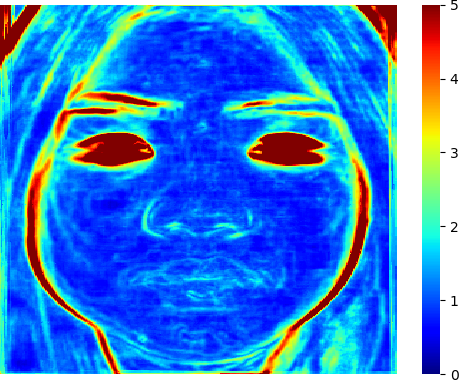}
		\label{fig:kp-vis}}
	\centering
	\subfigure[Proposed Joint]{\includegraphics[width=0.3\linewidth]{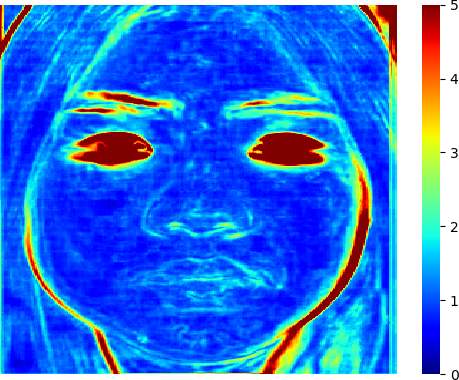}
		\label{fig:joint-vis}}
	\caption{Visualization of the Aligned Faces in Standard Deviation. The color map is within a scale of $5$ in standard deviation. Deeper red represents higher standard deviation value.}
	\label{fig:alignment-vis}
\end{figure}

Fig. \ref{fig:alignment-vis} visualizes the improvements compared to the landmark-only face alignment. The test video is selected from the HKBU-Mars V2 dataset \cite{liu2016hkbu-mars-v2}. 200 frames are aligned using landmark-only, keypoint-only, and the proposed face alignment method. 
The region outside the template after the projection is tailored so that all the aligned faces have the same size. For each method, all the aligned faces are overlaid and the heatmap for pixel-wise standard deviations is visualized. It can be seen that the proposed face alignment improves the precision significantly, \eg, the variations of the aligned faces are greatly reduced. The proposed face alignment method lays a solid foundation for the subsequent rPPG signal extraction.


\begin{figure*}[!t]
	\centering
	\includegraphics[width=7in]{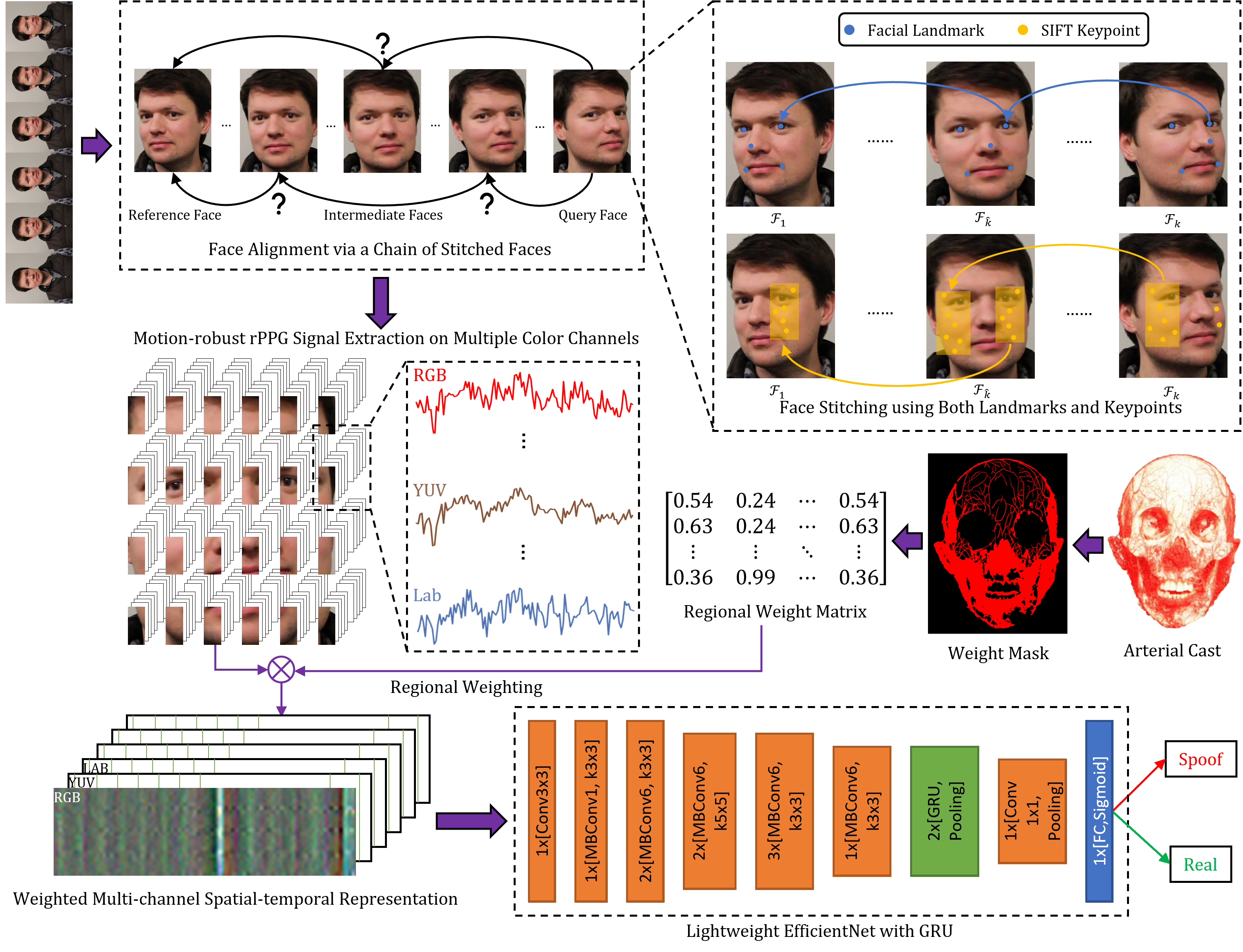}
	\caption{Overall diagram of the proposed Vascular-weighted Motion-robust rPPG (VMrPPG). Four building blocks are presented: 1) Face alignment via image stitching; 2) rPPG signal extraction from multiple color spaces; 3) Spatial-temporal representation weighted using the density of blood vessels; and 4) A customized EfficientNet with a GRU. The photo of arterial cast originates from \cite{sun2018colour}.}
	\label{fig:overview}
\end{figure*}

\section{Proposed rPPG-based Face Anti-spoofing}
\subsection{Overview of Proposed Framework}
rPPG signals are often weak and noisy due to head motions, alignment errors, and illumination variations. A set of techniques are developed in this paper to enhance the rPPG signals. In the previous section, a face alignment method via both landmarks and keypoints is proposed to robustly and precisely align the faces at the pixel-wise level. To focus on the ROIs with rich blood vessels, a novel signal weighting mechanism is proposed, where the weight for each ROI is determined by the density of blood vessels within the ROI. To exploit different patterns of rPPG signals in various color spaces, rPPG signals from multiple color spaces are combined to form the spatial-temporal representation. To learn a generalized model for 3D mask spoofing detection, a lightweight EfficientNet with a GRU is proposed by utilizing the compound scaling mechanism \cite{tan2019efficientnet}, which provides a wide adaption ability of modeling the rPPG signals in various applications.

The overall framework of the proposed Vascular-weighted Motion-robust rPPG is shown in Fig. \ref{fig:overview}, which includes four main building blocks. 1) Robust face alignment. Facial landmarks may not be stably detected to a pixel-wise level, but can be detected in most frames. In contrast, a few SIFT keypoints can be matched over a large pose difference, but they can be detected stably at a pixel level. The proposed face alignment method makes good use of both keypoints and landmarks to extract stable rPPG signals from the face video. 2) rPPG extraction from multiple color spaces. The rPPG signals are extracted from multiple ROIs 
to take account of their phase differences in different regions of a face. The extracted signals from RGB, YUV, and Lab color spaces are combined to form a consolidated signal representation. 3) Vascular-weighted spatial-temporal representation. The spatial-temporal representation encodes rPPG signals of different ROIs embedding the phase differences and the magnitude variations. The vascular-weighting mechanism weighs the encoded signals based on the blood vessel distribution from anatomical atlas to highlight the regions with rich blood vessels. 4) Classification using a customized EfficientNet with a GRU. The proposed classifier makes use of the EfficientNet blocks in spatial feature learning and the GRU in temporal feature learning.

\subsection{Vascular-weighted Spatial-temporal Representation}
\label{weights}
The faces in a video are first aligned using the proposed face stitching method in Section \ref{sec:joint-align}. 
The rPPG signals of the predefined ROIs in each frame are calculated as the average pixel values of a color channel as in \cite{niu2020rhythmnet}. The extracted rPPG signals contain various types of ineluctable noise from head micro-motions, alignment errors, illumination variations, etc. To filter the noise outside the heart rate, a bandpass filter is applied with the cutoff frequency at 0.85Hz and 3.5Hz, following the range of normal heart rates.

The filtered signal may still contain noise. The phase information is then utilized to distinguish the noise and the real signal. The blood flows out of the heart at a constant speed and reaches each ROI at a certain time, leaving a peak on the rPPG signal. As the distance from the heart to each ROI varies, the peak arrives at different times. With the same frequency, the rPPG signal in one ROI has a phase difference from those in other ROIs, while the phase of the noise is random. This unique phase information can be modeled as a robust liveness clue. To capture the phase pattern, the extracted signals from different ROIs are stacked to construct an image-like spatial-temporal representation, similarly as in~\cite{niu2020rhythmnet}. 
Formally, denote the rPPG signals from color channel $c$ as:
\begin{equation}
    \bm{S}^{c}=[\bm{r}^{c}_{1},\bm{r}^{c}_{2},...,\bm{r}^{c}_{M}]^\mathsf{T}.
\end{equation}
where $M$ refers to the number of ROIs. The rPPG signal in each ROI $\bm{r}^{c}_{i}$ is a sequence of the average pixel values of ROIs in color channel $c$, \ie, $\bm{r}^{c}_{i}=[r^{c}_{i,1},r^{c}_{i,2},...,r^{c}_{i,N}]$, where $r^{c}_{i,j}$ indicates the average pixel value of the $i$-th ROI of the $j$-th frame. 
The phase information is well embedded in the spatial-temporal representation without explicitly extracting it.


The magnitudes of rPPG signals from all ROIs are also important for spoofing detection. The rPPG signal originates from the periodical contraction of the facial blood vessels and presents periodical color changes on face skin \cite{wang2017algorithmic}. Larger density of blood vessels beneath a ROI results in larger magnitude of color changes. Thus, the ROIs with richer blood vessels should have stronger rPPG signals. To highlight the rPPG signals in the region with rich blood vessels, a signal weighting mechanism is designed to assign the weight according to the density of the blood vessels. 

\begin{figure}[htb]
	\centering
	\subfigure[Arterial cast image of head.]{\includegraphics[width=1.5in]{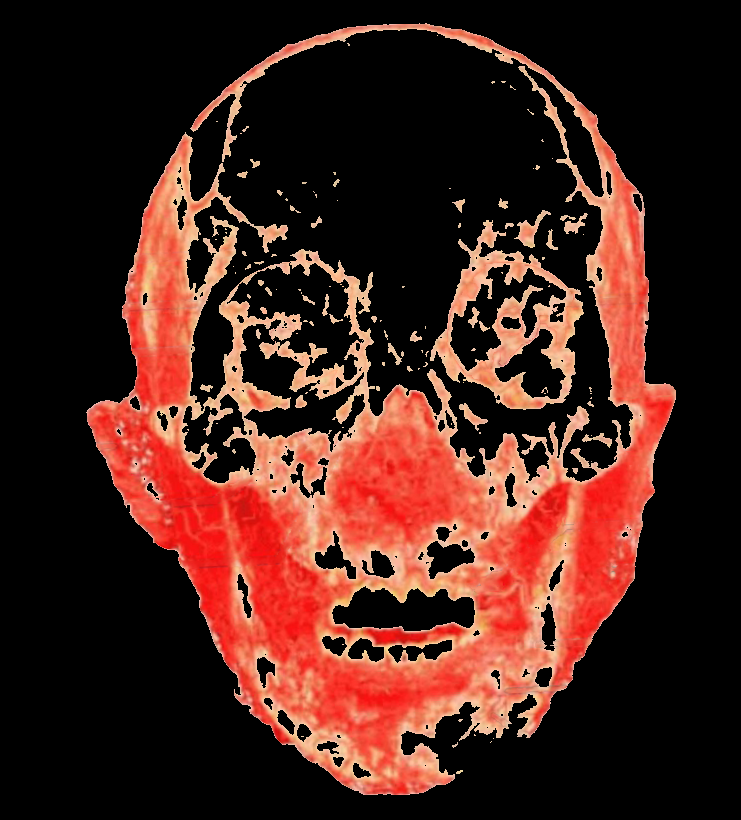}
		\label{fig:blood-vessel-dis}}
	\centering
	\subfigure[rPPG signals in a face \cite{liu2021multi}]{\includegraphics[width=1.5in]{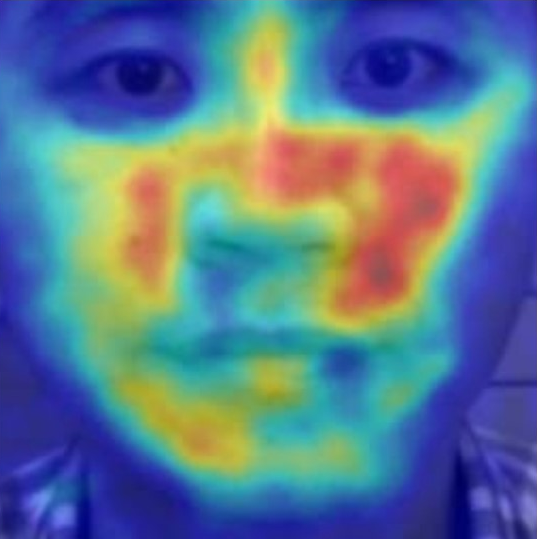}
		\label{fig:blood-vessel-hm}}
	\caption{Biology foundation of rPPG signals. The periodical blood vessel contraction results in the regular color changes of the skin. 
	The cheeks and mandibles have rich blood vessels and hence strong rPPG signals. These regions are hence assigned larger weights as in Eqn. (\ref{eqn:weighting}).}
	\label{fig:blood-vessel}
\end{figure}

An arterial cast of head is utilized to estimate the density of blood vessels in each ROI. To map the 3D arterial cast to 2D face image, the frontal view image is taken first and 5 landmarks (2 eye centers, 1 nose tip, and 2 mouth corners) are manually labeled. The skin area containing few blood vessels in the frontal view image are cropped. An affine transform is then applied using the 5 landmarks. One transformed arterial cast image is shown in Fig. \ref{fig:blood-vessel-dis}, which suggests that the cheeks and mandibles have the richest blood vessels while the forehead lacks rich blood vessels. It matches the rPPG signals measured in \cite{liu2021multi}, which uses the normalized per-pixel standard deviation to represent the rPPG signal in each pixel of a frontal human face. As only the skin covering dense blood vessels presents visible color changes, the weights are estimated proportional to the area of such skin. Formally, denote the weights at color channel $c$ as $\bm{w}^{c}=\{w^{c}_{1},w^{c}_{2},...,w^{c}_{M}\}$. 
The weighted rPPG signals $\hat{\bm{S}}^{c}$ are represented as:
\begin{equation}
	\hat{\bm{S}}^{c}=[w^{c}_{1}\bm{r}^{c}_{1},w^{c}_{2}\bm{r}^{c}_{2},...,w^{c}_{M}\bm{r}^{c}_{M}]^\mathsf{T}.
	\label{eqn:weighting}
\end{equation}

\subsection{Multi-color-space rPPG Representation}
rPPG signals represent the color changes of human face~\cite{wang2017algorithmic}, while the color changes are affected by devices, illumination conditions, body conditions, etc. To enhance the generalization ability under cross-domain scenarios, the rPPG signals are normalized in the range of $[0,1]$. 
Recent study \cite{niu2020rhythmnet} shows that the rPPG signals in different color spaces present different characteristics. For example, the rPPG signals in the blue channel is weaker as more blue light is absorbed by human skin. 
To construct a robust signal representation, the unique characteristics of rPPG signals in RGB, YUV, and Lab color spaces are jointly utilized in this paper.

\subsection{Customized EfficientNet with GRU}
The proposed representation contains temporal changes of rPPG signals, and spatial correlations of rPPG signals from different ROIs. 
To extract the discriminant features from the proposed representation, the building blocks of EfficientNet \cite{tan2019efficientnet} are utilized to make use of the power of EfficientNet in image classification \cite{tan2019efficientnet}. As the number of subjects in face-spoofing datasets \cite{erdogmus20133dmad,liu2016hkbu-mars-v2} is usually small, deeper networks may easily lead to overfit. To address this problem, the building blocks of EfficientNet are designed following the compound scaling mechanism but with a smaller size compared to the baseline architecture EfficientNet-B0 \cite{tan2019efficientnet}.

It should be noted that the horizontal axis is the proposed 2D spatial-temporal signal representation is the time axis. 
To obtain the liveness clues from both spatial and temporal dimension, a lightweight EfficientNet with GRU is designed to combine building blocks of EfficientNet with a Gated Recurrent Unit (GRU) \cite{comas2021turnip}. The EfficientNet building blocks are designed to extract the spatial-temporal features at multiple scales, while the GRU is designed to explicitly model the temporal relations.  
More specifically, the Gated Recurrent Unit is applied at the end of EfficientNet, taking one column of features as the input followed by successive columns along the time axis. The GRU is then designed to learn the temporal correlations between rPPG signals at successive time instances. 
The detailed network architecture is shown in Table~\ref{efficientnet}.
\begin{table}[htb]
    \caption{The architecture of the proposed network.}
	\begin{center}
		\begin{tabular}{c|c|c|c|c}
			\hline
			\small{Stage} & \small{Operator} & \small{Resolution} & \small{Channels} & \small{Layers} \\
			$i$ & $\mathcal{O}_{i}$ & $H_{i}\times W_{i}$ & $C_{i}$ & $L_{i}$ \\
			\hline
			1 & \small{Conv3x3} & $24\times120$ & $64$ & $1$ \\
			2 & \small{MBConv1,k3x3} & $12\times60$ & $24$ & $1$ \\
			3 & \small{MBConv6,k3x3} & $12\times60$ & $48$ & $2$ \\
			4 & \small{MBConv6,k5x5} & $6\times30$ & $80$ & $2$ \\
			5 & \small{MBConv6,k3x3} & $6\times30$ & $160$ & $3$ \\
			6 & \small{MBConv6,k3x3} & $3\times15$ & $224$ & $1$ \\
			7 & \small{GRU,Pooling} & $1\times15$ & $448$ & $2$ \\
			8 & \footnotesize{Conv1x1,Pooling,FC} & $1\times15$ & $448$ & $1$ \\
			\hline
		\end{tabular}
	\end{center}
	\label{efficientnet}
\end{table}

`MBConv' refers to mobile inverted bottleneck \cite{sandler2018mobilenetv2} with squeeze-and-excitation optimization \cite{hu2018squeeze}. The spatial-temporal representation has 18 channels: RGB, YUV, Lab, and their corresponding normalized channels. The proposed network outputs two confidence rates for live and spoof decisions. 

The loss function is defined as the cross entropy between the ground-truth and the prediction labels. Denote $\bm{u}_{i,j}$ as the confidence scores of the $i$-th batch of samples belonging to the $j$-th class and $\bm{y}_{i,j}$ as their ground-truth labels. The loss function of $N$ batches for $M$ classes is calculated as:
\begin{equation}
	\mathcal{L}=-\sum_{i=1}^{N}\sum_{j=1}^{M}\bm{y}_{i,j}\log\bm{u}_{i,j}+(1-\bm{y}_{i,j})\log(1-\bm{u}_{i,j}).
\end{equation}

The superiority of EfficientNet mainly lies in the compound scaling mechanism, which can concentrate more on relevant regions and preserve more object details flexibly, depending on the size of training data. For rPPG-based face anti-spoofing, especially for the cross-dataset evaluation, the sizes of training data vary significantly. The compound-scaling mechanism could help handle the variety of application scenarios.

\section{Experimental Results}


\subsection{Experimental Settings}
\subsubsection{Benchmark Datasets}
Four datasets are used for evaluating 3D mask attack detection: 3DMAD \cite{erdogmus20133dmad}, CSMAD \cite{bhattacharjee2018spoofing}, HKBU-Mars V1+ \cite{liu2018remote}, and HKBU-Mars V2 \cite{liu2016hkbu-mars-v2}, where the last two are collected under a similar protocol, but with no common video. And the Idiap Replay Attack dataset \cite{chingovska2012effectiveness} is used to detect printed photo attacks and video replay attacks.

\noindent \textbf{3DMAD} dataset  \cite{erdogmus20133dmad} consists of 2 sessions of videos on genuine faces and 1 session of videos on 3D mask attacks of 17 subjects. The used 3D masks are from Thatsmyface and the subjects vary in race, age, and gender. Each subject has 5 10-second videos in RGB and 5 10-second videos in depth. For a fair comparison, only the RGB videos are used. There are 300 frames of $640\times480$ pixels for each video. This dataset is collected under the indoor environment with well-controlled illumination conditions. 

\noindent \textbf{CSMAD} (Custom Silicone Mask Attack Dataset) \cite{bhattacharjee2018spoofing} was collected from 14 subjects and 6 high-quality 3D masks. The dataset consists of 87 genuine videos and 159 attack videos, in which the 3D masks are worn on subjects' faces or mounted on an appropriate stand. The frame rate is approximately 30 frames per second (FPS), lasting from 5 to 12 seconds. Four lighting conditions are adopted, including flourscent ceiling light, left halogen lamp illuminating, right halogen lamp illuminating, and halogen lamp illuminating from both sides. Each video contains frames in visible light recorded by Intel RealSense SR300 and near-infrared images recorded by Seel Thermal Compact Pro. In this paper, only the visible-light videos are employed for evaluation.

\noindent \textbf{HKBU-Mars V1+} dataset \cite{liu2018remote} has 2 sessions of genuine attempts and 1 session of 3D mask attacks. It only contains RGB videos. Each session includes 60 10-second videos for 12 genuine subjects or 3D masks. The subjects vary in age and gender. Due to the privacy issue, the public version eliminate one subject's videos on sessions of genuine attempts. Six masks are made by Thatsmyface and the other two are made by REAL-f. The videos are recorded via Logeitech C920 web-camera with the resolution of $1280\times720$ pixels at 25 FPS, under controlled laboratory light conditions. 

\noindent \textbf{HKBU-Mars V2} dataset \cite{liu2016hkbu-mars-v2} is much larger and covers more real-world variations. It contains 1 session of real-face videos and 1 session of spoofing attack videos. Every session has 504 10-second videos from 12 subjects or 3D masks. Similar to HKBU-Mars V1+, the subjects vary in age and gender. Six masks are made by Thatsmyface and the other six are made by REAL-f. In general, the subject's frontal face is recorded, allowing natural facial expressions and some head movements. Compared to HKBU-Mars V1+, this version introduces three more variations. 1) Multiple devices are used 
with FPSs ranging from $14$ to $50$ and resolutions ranging from $640\times480$ to $1920\times1080$ pixels. 2) The devices are either fixed on tripods or handheld, resulting in larger motions. 3) There are various lighting conditions, including room light, low light, bright light, warm light, side light and up side light.

\noindent \textbf{Idiap Replay Attack Dataset} \cite{chingovska2012effectiveness} consists of 1300 video clips of 50 subjects performing real attempts, printed photo attacks, and video replay attacks. 
The videos of real attempts are generated by users attempting to access a laptop through a Macbook built-in webcam while the spoofing attacks are performed by displaying a photo or video recording of the same user for at least 9 seconds. All videos are of the resolution of $320\times 240$ at $25$ FPS. Two lighting conditions are employed, the controlled office light with homogeneous backgrounds and the adverse illumination with complex backgrounds.

\subsubsection{Compared Methods}
The following methods are selected for comparison. 

\noindent \textbf{Multi-Scale Local Binary Pattern (MS-LBP)} \cite{erdogmus2014spoofing} is the baseline method of nearly all 3D mask attack datasets. It extracts multi-scale LBP-histogram features of $833$ dimensions and utilizes a support vector machine (SVM) with a linear kernel as the classifier. 

\noindent \textbf{Color Texture Analysis (CTA)} \cite{boulkenafet2015face} extracts $434$-dimension LBP features on both HSV and YCbCr color spaces. The extracted features are then classified by an SVM with an RBF kernel. It is chosen for comparison because of its good generalization ability in detecting general face attacks. 

\noindent \textbf{FBNet-RGB} \cite{shen2019facebagnet} is a representative deep neural network, which ranks the second in the Multi-modal Face Anti-spoofing Attack Detection Challenge of CVPR 2019. The method extracts features from image patches using a sequence of residual blocks. 


\noindent \textbf{GrPPG} \cite{li2016generalized} utilizes the Power Spectral Density curve generated by the Fast Fourier Transform (FFT) as the feature representation and extracts rPPG signals from ROIs. The features are then classified by an SVM with a linear kernel.

\noindent \textbf{PPGSecure} \cite{nowara2017ppgsecure} extracts signals from both skin area and backgrounds to construct spectral features using the FFT. The features are then classified by an RBF SVM.

\noindent \textbf{LrPPG} \cite{liu20163d} extracts a Local rPPG Confidence Map using the ridge regression of rPPG signals from multiple ROIs, and generate a Local rPPG Correlation Model as the features. The generated features are then classified by an RBF SVM.

\noindent \textbf{CFrPPG} \cite{liu2018remote} extracts the correspondence feature of local rPPG signals after the Fast Fourier Transform. The correlation features are extracted based on the Power Spectral Density curve and classified by a linear SVM.

\noindent \textbf{TSrPPG} \cite{liu2020temporal} excavates the similarity features between rPPG signals from multiple face regions, and the dissimilarity features between face regions and background regions by measuring the correspondences of signal amplitudes, gradients, and phases. 

\noindent \textbf{MCCFrPPG} \cite{liu2021multi} extends the CFrPPG \cite{liu2018remote} by applying segments of Short-Time Fourier Transform (STFT) to obtain spectrogram features. The multi-channel rPPG correspondence features are then classified by a linear SVM.

\noindent \textbf{TransRPPG} \cite{yu2021transrppg} adopts a vision transformer to extract the temporal information from the rPPG signals. The multi-scale spatial-temporal maps on both facial skin and background regions are generated with the size $63\times300\times3$ and $15\times300\times3$. Two network branches of share-weight Transformer Layers are designed to learn the attentional features.

\noindent \textbf{PATRON} \cite{birla2022patron} separates the respiratory signal from the original rPPG signal as a new liveness cue. The similarity features are extracted from both respiratory signals and original rPPG signals, and then classified by an RBF SVM.

\noindent \textbf{SUNRISE} \cite{birla2022sunrise} considers the similarity features of multiple ROIs from both temporal representation and spectral representation of rPPG signals, and uses multiple RBF SVMs on signal fragments of different sizes for classification.

\noindent \textbf{LeTSrPPG} \cite{liu2022learning} extends the TSrPPG \cite{liu2020temporal} by tuning the ROI frames with a C(2+1)D neural network for higher quality rPPGs. The network is trained via minimizing an rPPG regression loss measuring the similarity between rPPG signals of face ROIs and the dissimilarity between rPPG signals of face ROIs and that of background ROIs.

\subsubsection{Implementation Details}
The latest SeetaFace V6 \cite{wu2017funnel} is used to detect faces and facial landmarks. To handle videos under extreme light conditions, histogram equalization is applied to improve the image contrast. As recommended in~\cite{lowe2004distinctive}, the threshold to choose the initial matched keypoints $\delta$ is set to 0.6 for the distance defined in Eqn. (\ref{eqn:dmi}) so that more than $70$ initially matched SIFT keypoints can be found in real time. The relative weight $\lambda$ in Eqn. (\ref{eqn:joint-dis}) is empirically set to $3$ to highlight the spatial similarity on face sequences. The acceptance rate $\alpha$ is set to $0.5$ to reserve sufficient and high-quality point matches for calculating affine transformation. 

The face is split into $24$ ROIs following Niu \etal's design~\cite{niu2020rhythmnet}, with $4$ rows and $6$ columns to reserve the integrity of mandible regions. To learn a unified pattern from videos with different FPSs, the extracted rPPG signals are normalized to $30$ FPS via cubic interpolation. To make better use of the information from the whole video and adapt to videos of different sizes, a video is cut into segments of $120$ frames, with an overlapping of $117$ frames. Having $9$ color channels and $9$ normalized color channels, the size of the proposed weighted spatial-temporal representation is $[24,120,18]$. The learning rate is set to $0.1$ initially with a decay to $10\%$ every $4$ epochs and a L2 regularization penalty of $5\times10^{-4}$. A decision can be made for each video segment whether it is a genuine attempt or a spoofing attack. The final decision is the majority vote of the results from all video segments.

\subsubsection{Evaluation Metrics}
The following evaluation metrics are reported.

\noindent \textbf{Equal Error Rate (EER)} refers to the value when the False Acceptance Rate (FAR) and the False Reject Rate (FRR) are equal.

\noindent \textbf{Half Total Error Rate (HTER)} is evaluated at a threshold for the EER on the development set, and it is calculated as $HTER=(FAR+FRR)/2$.

\noindent \textbf{BPCER@APCER=0.1} represents the BPCER value when APCER is $0.1$, where APCER (Attack Presentation Classification Error Rate) and BPCER (Bonafide Presentation Classification Error Rate) are similar to FAR and FRR, but the threshold for APCER=0.1 is determined using the development set and applied on the test set when calculating BPCER.

\noindent \textbf{BPCER@APCER=0.01} represents the BPCER value when APCER is $0.01$.


\noindent \textbf{Area Under the Curve (AUC)} indicates the area under the Receiver Operating Characteristic curve. 


\subsection{Experimental Results of Intra-dataset Evaluation}
\subsubsection{Experimental Results on the 3DMAD Dataset}
\label{sec:intra-3dmad}
For a fair comparison, the standard evaluation protocol, Leave-One-Out-Cross-Validation (LOOCV) \cite{liu2018remote,liu2021multi}, is used. In each fold, one subject is used as the test set, $8$ subjects are randomly chosen as the train set, and the remaining $8$ subjects are used as the development set. $20$ rounds of LOOCV are conducted to avoid coincidence and the average results over 20 rounds are reported. The comparisons to state-of-the-art methods on 3DMAD are shown in Table \ref{table:intra-3dmad}. We implemented and evaluated LeTSrPPG \cite{liu2022learning} on the 3DMAD dataset. The results of other compared methods are reported from their original~papers. 
\begin{table}[!t]
    \caption{Comparison on the 3DMAD dataset. The best scores are marked in \textbf{bold} and the second best in \underline{underline}. The rPPG-based methods are marked with $\star$ while appearance-based methods with $\blacktriangle$.}
	\begin{center}
	\resizebox{1.0\columnwidth}{!}{
		\begin{tabular}{|l|c|c|c|c|}
			\hline
			Method & HTER\_dev & HTER\_test & EER & AUC \\
			\hline
			$\blacktriangle$ MS-LBP \cite{erdogmus2014spoofing} & $\bm{1.25 \pm 1.90}$ & $\underline{4.22 \pm 10.30}$ & $2.66$ & $\bm{99.60}$ \\
			$\blacktriangle$ CTA \cite{boulkenafet2015face} & $2.78 \pm 3.60$ & $4.40 \pm 9.70$ & $4.24$ & $99.30$ \\
			$\blacktriangle$ FBNet-RGB \cite{shen2019facebagnet} & $3.91 \pm 2.40$ & $5.66 \pm 9.70$ & $5.54$ & $98.60$ \\
			$\star$ GrPPG \cite{li2016generalized} & $13.40 \pm 4.20$ & $13.20 \pm 13.20$ & $13.90$ & $92.60$ \\
			$\star$ PPGSecure \cite{nowara2017ppgsecure} & $15.20 \pm 4.40$ & $15.90 \pm 14.60$ & $15.80$ & $90.80$ \\
			$\star$ LrPPG \cite{liu20163d} & $9.06 \pm 4.40$ & $8.57 \pm 13.30$ & $8.88$ & $96.00$ \\
			$\star$ CFrPPG \cite{liu2018remote} & $5.95 \pm 3.30$ & $6.82 \pm 12.10$ & $6.94$ & $97.10$ \\
			$\star$ MCCFrPPG \cite{liu2021multi} & $4.42 \pm 2.30$ & $5.60 \pm 8.80$ & $5.01$ & $98.70$ \\
			$\star$ TransRPPG \cite{yu2021transrppg} & $ - $ & $ - $ & \underline{$2.38$} & $98.80$ \\
                {$\star$ LeTSrPPG \cite{liu2022learning}} & {$11.15 \pm 2.99$} & {$12.35 \pm 10.86$} & {$7.65$} & {$95.53$} \\
			\hline
			$\star$ Proposed VMrPPG & \underline{$1.34 \pm 1.51$} & $\bm{2.16 \pm 4.17}$ & $\bm{0.87}$ & \underline{$99.58$} \\
			\hline
		\end{tabular}}
	
	\end{center}
	\label{table:intra-3dmad}
\end{table}

\begin{table}[!t]
    \caption{Comparisons on the 3DMAD dataset for short-time observations. The rPPG-based methods designed specifically for short-time observation are marked with $\circ$.}
	\begin{center}
	\resizebox{1.0\columnwidth}{!}{
		\begin{tabular}{|l|c|c|c|c|}
			\hline
			Method & HTER\_dev & HTER\_test & EER & AUC \\
			\hline
			$\star$ GrPPG \cite{li2016generalized} & $34.10 \pm 5.70$ & $33.70 \pm 11.60$ & $38.30$ & $65.90$ \\
			$\star$ PPGSecure \cite{nowara2017ppgsecure} & $33.30 \pm 3.10$ & $33.00 \pm 8.10$ & $34.80$ & $69.40$ \\
			$\star$ LrPPG \cite{liu20163d} & $45.20 \pm 3.20$ & $44.80 \pm 8.80$ & $45.30$ & $55.70$ \\
			$\star$ CFrPPG \cite{liu2018remote} & $32.80 \pm 1.70$ & $32.70 \pm 7.40$ & $32.50$ & $70.80$ \\
			$\star$ TransRPPG \cite{yu2021transrppg} & $ 20.70 \pm 2.20 $ & $ 20.60 \pm 8.30 $ & $20.80$ & $84.50$ \\
                $\circ$ TSrPPG \cite{liu2020temporal} & $13.10 \pm 3.00$ & $13.40 \pm 11.20$ & $13.30$ & $93.80$ \\
                $\circ$ SUNRISE \cite{birla2022sunrise} & $ - $ & $ - $ & $12.50$ & $93.70$ \\
                $\circ$ LeTSrPPG \cite{liu2022learning} & \underline{$11.5 \pm 2.70$} & \underline{$ 11.80 \pm 8.60 $} & \underline{$11.90$} & \underline{$94.40$} \\
			\hline
			$\star$ Proposed VMrPPG & $\bm{3.67 \pm 1.33}$ & $\bm{5.00 \pm 7.06}$ & $\bm{4.41}$ & $\bm{98.09}$ \\
			\hline
		\end{tabular}
	}
	\end{center}
	\label{table:intra-3dmad-1s}
\end{table}

It is witnessed that the proposed approach reduces the HTER of the test set of the previously best performed rPPG-based method, MCCFrPPG \cite{liu2021multi}, from $5.60\%$ to $2.16\%$, and reduces the EER of the previously best performed method TransRPPG \cite{yu2021transrppg} from $2.38\%$ to $0.87\%$. In terms of the AUC, the proposed method increases the result of the previously best performed rPPG-based method, TransRPPG \cite{yu2021transrppg}, from $98.80\%$ to $99.58\%$.    
It is also worth to mention that the standard derivations of the HTER of test set for all evaluated methods are high. After checking the dataset, the genuine face videos for two subjects with aging face and swarthy skin are more likely to be wrongly classified as spoofing faces, which suggests that the current face spoofing detection methods are still greatly suffered from age and skin color variations.

TSrPPG \cite{liu2020temporal}, SUNRISE \cite{birla2022sunrise}, and LeTSrPPG \cite{liu2022learning} were originally designed for short-time observation scenarios. For a fair comparison, the proposed method is also evaluated on short-time observation (1 second) scenarios following the same protocol as in \cite{liu2022learning}. The results are summarized in Table \ref{table:intra-3dmad-1s}. All results of the compared methods are reported from their original papers. It can be seen that the proposed VMrPPG significantly outperforms all the compared methods on all four evaluation criteria. Specifically, compared with the previously best performed method, LeTSrPPG \cite{liu2022learning}, the proposed method reduces the EER from $11.90\%$ to $4.41\%$, and increases the AUC from $94.40\%$ to $98.09\%$.

\subsubsection{Experimental Results on the HKBU-Mars V1+ Dataset}
With fewer subjects than the 3DMAD dataset, training on this dataset requires higher generalization ability on the model. On this dataset, the penalty for L2 regularization is elevated to $5\times10^{-3}$ and the learning rate decays after $4$ epochs. 20 rounds of LOOCV are applied on the HKBU-Mars V1+ dataset. In each fold, one subject is used for testing, $5$ subjects are randomly selected for training, and the $6$ remaining subjects are used as the development set. The evaluation results are summarized in Table \ref{table:intra-hkbu-mars-v1}. 
It can be seen that the proposed VMrPPG achieves the best score on all four evaluation metrics, which outperforms the second best MCCFrPPG \cite{liu2021multi} by $1.37\%$,  $0.54\%$, $2.33\%$, and $0.03\%$ in terms of the HTER on the development set, the HTER on the test set, the EER, and the AUC, respectively. Different from the results on the 3DMAD dataset, the appearance-based methods present a sharp performance degradation on the HKBU-Mars V1+ dataset while rPPG-based methods remain their good performance. This phenomenon suggests that, under more scenario variations, the rPPG signals tend to be more robust than the appearance features.

We have also conducted the comparison experiments for short-time observations on the HKBU-Mars V1+ dataset, following the experimental settings in \cite{liu2022learning}. The results are summarized in Table~\ref{table:intra-hkbu-mars-v1-1s}. Similar to the results on the 3DMAD dataset, the proposed VMrPPG ranks the first on all four evaluation criteria, which suggests that the proposed VMrPPG significantly outperforms the state-of-the-art methods for both long-time observations and short-time observations.

\begin{table}[!t]
    \caption{Comparison on the HKBU-Mars V1+ dataset.}
	\begin{center}
	\resizebox{1.0\columnwidth}{!}{
		\begin{tabular}{|l|c|c|c|c|}
			\hline
			Method & HTER\_dev & HTER\_test & EER & AUC \\
			\hline
			$\blacktriangle$ MS-LBP \cite{erdogmus2014spoofing} & $20.50 \pm 8.90$ & $24.00 \pm 25.60$ & $24.80$ & $84.50$ \\
			$\blacktriangle$ CTA \cite{boulkenafet2015face} & $22.40 \pm 10.40$ & $23.40 \pm 20.50$ & $23.30$ & $81.90$ \\
			$\blacktriangle$ FBNet-RGB \cite{shen2019facebagnet} & $35.00 \pm 11.30$ & $36.10 \pm 26.00$ & $36.40$ & $67.30$ \\
			$\star$ GrPPG \cite{li2016generalized} & $15.40 \pm 6.70$ & $15.40 \pm 20.40$ & $16.20$ & $89.30$ \\
			$\star$ PPGSecure \cite{nowara2017ppgsecure} & $14.20 \pm 5.80$ & $15.60 \pm 16.40$ & $17.40$ & $90.70$ \\
			$\star$ LrPPG \cite{liu20163d} & $8.43 \pm 2.90$ & $8.67 \pm 8.80$ & $8.94$ & $97.10$ \\
			$\star$ CFrPPG \cite{liu2018remote} & $3.24 \pm 2.00$ & $4.10 \pm 4.90$ & $4.00$ & $99.30$ \\
			$\star$ MCCFrPPG \cite{liu2021multi} & \underline{$2.85 \pm 1.80$} & \underline{$3.38 \pm 4.80$} & \underline{$3.10$} & \underline{$99.70$} \\
                {$\star$ PATRON \cite{birla2022patron}} & {$ - $} & {$ - $} & {$14.70$} & {$87.80$} \\
                {$\star$ LeTSrPPG \cite{liu2022learning}} & {$7.64 \pm 4.37$} & {$4.09 \pm 7.33$} & {$7.27$} & {$95.64$} \\
			\hline
			$\star$ Proposed VMrPPG & $\bm{1.48 \pm 1.66}$ & $\bm{2.84 \pm 5.01}$ & $\bm{0.77}$ & $\bm{99.73}$ \\
			\hline
		\end{tabular}
	}
	\end{center}
	\label{table:intra-hkbu-mars-v1}
\end{table}

\begin{table}[!t]
    \caption{Comparisons on the HKBU-Mars V1+ dataset for short-time observation.}
	\begin{center}
	\resizebox{1.0\columnwidth}{!}{
		\begin{tabular}{|l|c|c|c|c|}
			\hline
			Method & HTER\_dev & HTER\_test & EER & AUC \\
			\hline
			$\star$ GrPPG \cite{li2016generalized} & $29.20 \pm 4.70$ & $29.10 \pm 9.70$ & $33.80$ & $72.00$ \\
			$\star$ PPGSecure \cite{nowara2017ppgsecure} & $42.40 \pm 2.10$ & $42.90 \pm 5.80$ & $43.00$ & $59.30$ \\
			$\star$ LrPPG \cite{liu20163d} & $45.30 \pm 3.70$ & $45.10 \pm 12.00$ & $45.30$ & $56.20$ \\
			$\star$ CFrPPG \cite{liu2018remote} & $41.60 \pm 3.30$ & $43.10 \pm 5.60$ & $42.00$ & $60.80$ \\
			$\star$ TransRPPG \cite{yu2021transrppg} & $ 32.90 \pm 2.80 $ & $ 32.70 \pm 6.40 $ & $33.10$ & $72.00$ \\
                $\circ$ TSrPPG \cite{liu2020temporal} & $21.50 \pm 2.60$ & $22.30 \pm 8.80$ & $22.00$ & $85.20$ \\
                $\circ$ SUNRISE \cite{birla2022sunrise} & $ - $ & $ - $ & \underline{$14.80$} & $86.80$ \\
                $\circ$ LeTSrPPG \cite{liu2022learning} & \underline{$15.30 \pm 2.20$} & \underline{$ 15.80 \pm 6.50 $} & $15.70$ & \underline{$91.50$} \\
			\hline
			$\star$ Proposed VMrPPG & $\bm{5.77 \pm 1.92}$ & $\bm{3.43 \pm 4.91}$ & $\bm{5.54}$ & $\bm{97.76}$ \\
			\hline
		\end{tabular}
	}
	\end{center}
	\label{table:intra-hkbu-mars-v1-1s}
\end{table}

\subsubsection{Experimental Results on the HKBU-Mars V2 Dataset}
\label{sec:intra-hkbuv2}
The standard evaluation protocol, 20 rounds of LOOCV, is applied on the HKBU-Mars V2 dataset, with the train-development-test subject-amount split as $(5,6,1)$. 
The results are summarized in Table \ref{table:intra-hkbu-mars-v2}. The results of other compared methods are obtained from \cite{liu2021multi}. 
The proposed VMrPPG consistently outperforms all the compared methods for all three evaluation metrics. 
Compared to the state-of-the-art method, MCCFrPPG \cite{liu2021multi}, the performance gains of the proposed method are $1.26\%$, $0.26\%$, and $10.26\%$ in terms of the EER, the AUC, and BPCER@APCER=0.01, respectively. The superiority of the rPPG-based methods over the appearance-based methods are more distinct, which exhibits the discriminant power of rPPG-based methods to overcome the variations of backgrounds, illumination conditions, and camera sensors. 

\begin{table}[!t]
    \caption{Comparison on the HKBU-Mars V2 dataset.}
	\begin{center}
		\begin{tabular}{|l|c|c|c|c|}
			\hline
			Method & EER & AUC & BPCER@APCER=0.01\\
			\hline
			$\blacktriangle$ MS-LBP \cite{erdogmus2014spoofing} & $22.50$ & $85.80$ & $95.10$ \\
			$\blacktriangle$ CTA \cite{boulkenafet2015face} & $23.00$ & $82.30$ & $89.20$ \\
			$\star$ GrPPG \cite{li2016generalized} & $16.40$ & $89.40$ & $32.90$ \\
			$\star$ LrPPG \cite{liu20163d} & $9.07$ & $97.00$ & $38.90$ \\
			$\star$ CFrPPG \cite{liu2018remote} & \underline{$4.04$} & \underline{$99.30$} & \underline{$17.80$} \\
			$\star$ TransRPPG \cite{yu2021transrppg} & $8.47$ & $96.80$ & $29.80$ \\
			\hline
			$\star$ Proposed VMrPPG & $\bm{2.78}$ & $\bm{99.56}$ & $\bm{7.54}$ \\
			\hline
		\end{tabular}
	\end{center}
	\label{table:intra-hkbu-mars-v2}
\end{table}

\begin{table}[!t]
    \caption{Comparison on the CSMAD dataset under the Leave-Half-Out-for-Training protocol~\cite{liu2021multi}.}
	\begin{center}
	\resizebox{1.0\columnwidth}{!}{
		\begin{tabular}{|l|c|c|c|c|c|}
			\hline
			Method & HTER\_test & EER & AUC & \makecell[l]{BPCER@\\APCER=0.1} & \makecell[l]{BPCER@\\APCER=0.01} \\
			\hline
			$\blacktriangle$ MS-LBP \cite{erdogmus2014spoofing} & \underline{$8.36 \pm 4.20$} & \underline{$9.28$} & \underline{$96.20$} & \underline{$7.65$} & $55.80$ \\
			$\blacktriangle$ CTA \cite{boulkenafet2015face} & $11.10 \pm 4.60$ & $12.90$ & $94.70$ & $16.60$ & $48.70$ \\
			$\blacktriangle$ FBNet-RGB \cite{shen2019facebagnet} & $39.30 \pm 4.20$ & $40.30$ & $63.80$ & $83.10$ & $97.90$ \\
			$\star$ GrPPG \cite{li2016generalized} & $35.70 \pm 2.80$ & $37.20$ & $70.20$ & $68.60$ & $91.70$ \\
			$\star$ PPGSecure \cite{nowara2017ppgsecure} & $21.90 \pm 5.70$ & $23.30$ & $83.60$ & $32.60$ & $56.40$ \\
			$\star$ LrPPG \cite{liu20163d} & $19.10 \pm 5.00$ & $19.60$ & $87.60$ & $32.90$ & $82.50$ \\
			$\star$ CFrPPG \cite{liu2018remote} & $12.50 \pm 3.00$ & $12.20$ & $93.80$ & $15.70$ & $59.70$ \\
			$\star$ MCCFrPPG \cite{liu2021multi} & $10.30 \pm 2.90$ & $10.70$ & $94.60$ & $11.30$ & \underline{$34.00$} \\
			\hline
			$\star$ Proposed VMrPPG & $\bm{6.91 \pm 2.40}$ & $\bm{8.00}$ & $\bm{97.36}$ & $\bm{7.15}$ & $\bm{32.77}$ \\
			\hline
		\end{tabular}
	}
	\end{center}
	\label{table:intra-csmad}
\end{table}

\subsubsection{Experimental Results on the CSMAD Dataset}

Following \cite{liu2021multi}, the Leave-Half-Out-for-Training protocol is adopted on the CSMAD dataset, which sets aside 7 subjects of genuine faces and 3 subjects of mask attacks for training, and leaves the rest for testing. The results on the CSMAD dataset are summarized in Table~\ref{table:intra-csmad}. The proposed VMrPPG performs best on all evaluation metrics. Compared with the second-best rPPG-based method, MCCFrPPG \cite{liu2021multi}, it reduces the HTER\_test by $3.39\%$, the EER by $2.70\%$, the BPCER@APCER=0.1 by $4.15\%$, and the BPCER@APCER=0.01 by $1.23\%$, and boosts the AUC by $2.76\%$.  
It is also noted that the rPPG-based methods perform poorer on this dataset than the other 3 datasets. After checking the failure cases, more than 70\% of them are genuine faces with side lighting that are incorrectly classified as masked faces. The results show that the rPPG signals are still strongly affected by the illumination conditions. 

From the intra-dataset evaluation on all four dataset, it can be concluded that the proposed VMrPPG consistently and significantly outperforms all the state-of-the-art face anti-spoofing methods based on rPPG signals. 

\begin{table*}
    \setlength\tabcolsep{3pt}
    \caption{Comparison on the 3DMAD, CSMAD, and HKBU-Mars V1+ datasets under the cross-dataset evaluation. "A$\rightarrow$B" refers to training on dataset A and testing on dataset B.}
    \begin{center}
    \resizebox{2.0\columnwidth}{!}{
        \begin{tabular}{|l|l|c|c|c|c|c|c|c|c|}
        \hline
        \multirow{2}*{Settings} & \multirow{2}{*}{Methods} & \multicolumn{4}{c|}{A $\rightarrow$ B}  & \multicolumn{4}{c|}{B $\rightarrow$ A}  \\
        \cline{3-10}
        & & HTER\_test & AUC & \minitab[c]{BPCER@\\APCER=0.1} & \minitab[c]{BPCER@\\APCER=0.01} & HTER\_test & AUC & \minitab[c]{BPCER@\\APCER=0.1} & \minitab[c]{BPCER@\\APCER=0.01} \\
        \hline
        \multirow{9}{*}{\minitab[l]{A: 3DMAD\\B: HKBU-Mars V1+}} & $\blacktriangle$ MS-LBP \cite{erdogmus2014spoofing} & $36.80 \pm 2.90$ & $60.70$ & $87.50$ & $97.00$ & $41.30 \pm 14.00$ & $62.20$ & $89.20$ & $99.50$ \\
        & $\blacktriangle$ CTA \cite{boulkenafet2015face} & $71.80 \pm 2.10$ & $45.90$ & $96.80$ & $99.30$ & $55.70 \pm 8.70$ & $48.60$ & $89.90$ & $97.40$ \\
        & $\blacktriangle$ FBNet-RGB \cite{shen2019facebagnet} & $34.00 \pm 1.40$ & $73.60$ & $65.70$ & $97.80$ & $12.30 \pm 10.60$ & $89.60$ & $26.80$ & $66.80$ \\
        & $\star$ GrPPG \cite{li2016generalized} & $35.90 \pm 4.50$ & $67.20$ & $75.80$ & $97.60$ & $36.50 \pm 6.80$ & $66.50$ & $86.30$ & $98.60$ \\
        & $\star$ PPGSecure \cite{nowara2017ppgsecure} & $14.40 \pm 1.40$ & $91.80$ & $16.90$ & $25.80$ & $19.10 \pm 2.30$ & $87.20$ & $26.20$ & $45.00$ \\
        & $\star$ LrPPG \cite{liu20163d} & $4.46 \pm 0.90$ & $98.90$ & $1.33$ & $31.20$ & $8.46 \pm 0.30$ & $95.30$ & $8.79$ & $17.00$ \\
        & $\star$ CFrPPG \cite{liu2018remote} & $4.23 \pm 0.30$ & $99.00$ & $2.83$ & $19.90$ & $4.81 \pm 0.40$ & $98.10$ & $4.44$ & $14.30$ \\
        & $\star$ MCCFrPPG \cite{liu2021multi} & \underline{$3.46 \pm 0.60$} & \underline{$99.60$} & $\bm{0.25}$ & \underline{$7.63$} & \underline{$4.78 \pm 0.80$} & \underline{$98.50$} & \underline{$4.00$} & \underline{$8.59$} \\
        & $\star$ Proposed VMrPPG & $\bm{2.21 \pm 0.79}$ & $\bm{99.73}$ & \underline{$0.50$} & $\bm{6.18}$ & $\bm{3.47 \pm 0.98}$ & $\bm{99.04}$ & $\bm{2.84}$ & $\bm{8.37}$ \\
        \hline
        {\multirow{9}{*}{\minitab[l]{A: 3DMAD\\B: CSMAD}}} & {$\blacktriangle$ MS-LBP \cite{erdogmus2014spoofing}} & {$50.60 \pm 5.60$} & {$49.50$} & {$89.90$} & {$98.10$} & {$42.70 \pm 6.40$} & {$58.20$} & {$83.90$} & {$98.90$} \\
       & {$\blacktriangle$ CTA \cite{boulkenafet2015face}} & {$48.90 \pm 5.80$} & {$50.70$} & {$88.60$} & {$98.10$} & {$58.40 \pm 7.80$} & {$46.50$} & {$93.60$} & {$99.30$} \\
       & {$\blacktriangle$ FBNet-RGB \cite{shen2019facebagnet}} & {$46.30 \pm 2.30$} & {$56.60$} & {$80.80$} & {$99.00$} & {$50.20 \pm 18.10$} & {$52.50$} & {$87.50$} & {$98.50$} \\
       & {$\star$ GrPPG \cite{li2016generalized}} & {$43.60 \pm 3.70$} & {$52.70$} & {$85.50$} & {$96.50$} & {$50.00 \pm 0.00$} & {$50.00$} & {$90.00$} & {$99.00$} \\
       & {$\star$ PPGSecure \cite{nowara2017ppgsecure}} & {$43.60 \pm 1.50$} & {$60.70$} & {$87.40$} & {$98.60$} & {$24.80 \pm 11.90$} & {$77.20$} & {$46.50$} & {$64.60$} \\
       & {$\star$ LrPPG \cite{liu20163d}} & {$40.50 \pm 2.60$} & {$67.00$} & {$63.50$} & {$82.30$} & {$17.00 \pm 7.20$} & {$84.70$} & {$46.30$} & {$99.90$} \\
       & {$\star$ CFrPPG \cite{liu2018remote}} & {$22.70 \pm 0.60$} & {$82.60$} & {$51.10$} & {$89.10$} & {$6.37 \pm 1.00$} & {$96.30$} & {$8.26$} & {$16.70$} \\
       & {$\star$ MCCFrPPG \cite{liu2021multi}} & {\underline{$9.98 \pm 0.40$}} & {$\bm{95.70}$} & {\underline{$10.90$}} & {\underline{$46.90$}} & {\underline{$3.71 \pm 0.80$}} & {\underline{$98.60$}} & {\underline{$3.47$}} & {\underline{$8.09$}} \\
       & {$\star$ Proposed VMrPPG} & {$\bm{9.18 \pm 1.04}$} & {\underline{$95.40$}} & {$\bm{10.57}$} & {$\bm{38.51}$} & {$\bm{2.94 \pm 1.33}$} & {$\bm{99.41}$} & {$\bm{2.23}$} & {$\bm{7.36}$}  \\
        \hline
        {\multirow{9}{*}{\minitab[l]{A: HKBU-Mars V1+\\B: CSMAD}}} & {$\blacktriangle$ MS-LBP \cite{erdogmus2014spoofing}} & {$42.30 \pm 3.20$} & {$52.30$} & {$85.40$} & {$97.90$} & {$45.00 \pm 5.80$} & {$54.80$} & {$87.10$} & {$99.00$} \\
       & {$\blacktriangle$ CTA \cite{boulkenafet2015face}} & {$53.60 \pm 5.00$} & {$48.80$} & {$90.30$} & {$98.70$} & {$37.80 \pm 4.80$} & {$61.50$} & {$80.20$} & {$96.20$} \\
       & {$\blacktriangle$ FBNet-RGB \cite{shen2019facebagnet}} & {$41.60 \pm 3.70$} & {$57.00$} & {$87.10$} & {$99.10$} & {$40.90 \pm 6.80$} & {$56.10$} & {$86.00$} & {$97.20$} \\
       & {$\star$ GrPPG \cite{li2016generalized}} & {$54.00 \pm 11.40$} & {$48.90$} & {$88.70$} & {$98.60$} & {$50.00 \pm 0.00$} & {$50.00$} & {$90.00$} & {$99.00$} \\
       & {$\star$ PPGSecure \cite{nowara2017ppgsecure}} & {$52.20 \pm 2.20$} & {$52.10$} & {$91.60$} & {$99.90$} & {$37.60 \pm 3.90$} & {$53.30$} & {$83.70$} & {$96.00$} \\
       & {$\star$ LrPPG \cite{liu20163d}} & {$40.40 \pm 2.90$} & {$65.40$} & {$65.10$} & {$81.50$} & {$13.80 \pm 8.00$} & {$88.60$} & {$39.20$} & {$98.10$} \\
       & {$\star$ CFrPPG \cite{liu2018remote}} & {$22.50 \pm 0.70$} & {$84.00$} & {$47.80$} & {$85.80$} & {\underline{$2.58 \pm 0.80$}} & {$99.30$} & {$1.29$} & {$17.30$} \\
       & {$\star$ MCCFrPPG \cite{liu2021multi}} & {\underline{$10.80 \pm 0.50$}} & {\underline{$95.30$}} & {\underline{$12.00$}} & {\underline{$40.90$}} & {$2.67 \pm 0.90$} & {\underline{$99.70$}} & {\underline{$0.75$}} & {\underline{$5.50$}} \\
       & {$\star$ Proposed VMrPPG} & {$\bm{9.44 \pm 1.10}$} & {$\bm{95.37}$} & {$\bm{10.75}$} & {$\bm{37.58}$} & {$\bm{1.93 \pm 1.05}$} & {$\bm{99.76}$} & {$\bm{0.59}$} & {$\bm{4.50}$} \\
            \hline  
        \end{tabular}
    }
    \end{center}
    \label{table:cross-3dmad-csmad-hkbuv1p}
\end{table*}

\subsection{Experimental Results of Cross-dataset Evaluation}
To evaluate the generalization ability of the proposed method on unseen scenarios, a set of cross-dataset evaluations are conducted following the same protocol as in MCCFrPPG \cite{liu2021multi}, in which the model is trained on Dataset A but evaluated on Dataset B (denoted as A $\rightarrow$ B). For the HKBU-Mars V1+ dataset, $6$ subjects are randomly selected as the train set and the remaining $6$ subjects are used as the development set. For the 3DMAD dataset, the train-development split is $(8,9)$. The CSMAD dataset treats 7 subjects for real attempts and 3 subjects for mask attacks for training respectively, and the rest as the development set. The cross-dataset evaluations are also repeated for $20$ rounds and the average results over $20$ rounds are reported. The results are summarized in Table \ref{table:cross-3dmad-csmad-hkbuv1p}. 

\subsubsection{3DMAD vs. HKBU-Mars V1+}
The results of the compared methods are obtained from \cite{liu2021multi}. 
For the setting of ``3DMAD $\rightarrow$ HKBU-Mars V1+", the proposed VMrPPG performs best on all three evaluation metrics, \ie, the HTER\_test of $2.21\%$, the AUC of $99.73\%$, and the BPCER@APCER=0.01 of $6.18\%$, which are better than the second-best method, MCCFrPPG \cite{liu2021multi}, by $1.25\%$, $0.13\%$, and $1.45\%$, respectively. The proposed VMrPPG ranks the second best for BPCER@APCER=0.1, which is only $0.25\%$ worse than MCCFrPPG \cite{liu2021multi}. For ``HKBU-Mars V1+ $\rightarrow$ 3DMAD", the proposed VMrPPG achieves the best performance in terms of all evaluation criteria. Compared to the previous best method MCCFrPPG \cite{liu2021multi}, the proposed method achieves a performance gain of $1.31\%$, $0.54\%$, $1.16\%$, and $0.22\%$ in terms of the HTER\_test, the AUC, the BPCER@APCER=0.1, and the BPCER@APCER=0.01, respectively. 

\subsubsection{3DMAD vs. CSMAD}
In this experiment, the proposed VMrPPG ranks the first on 7 evaluation criteria out of 8, as shown in Table \ref{table:cross-3dmad-csmad-hkbuv1p}. The only metric where the proposed method ranks the second is the AUC for ``3DMAD $\rightarrow$ CSMAD", which is $0.30\%$ lower than MCCFrPPG \cite{liu2021multi}. But the proposed VMrPPG outperforms MCCFrPPG \cite{liu2021multi} on all the other 7 evaluation criteria, 6 of which are significant.

\subsubsection{HKBU-Mars V1+ vs. CSMAD}
As shown in Table \ref{table:cross-3dmad-csmad-hkbuv1p}, the proposed VMrPPG ranks the first on all 8 evaluation criteria in this experiment. Compared to the second-best method MCCFrPPG \cite{liu2021multi}, the proposed VMrPPG reduces the HTER on the test set by $1.36\%$ for ``HKBU-Mars V1+ $\rightarrow$ CSMAD". For ``CSMAD $\rightarrow$ HKBU-Mars V1+", the proposed method reduce the HTER on the test set by $0.65\%$ compared to the second-best method CFrPPG \cite{liu2018remote}. Since the CSMAD dataset contains more illumination conditions than the 3DMAD or HKBU-Mars V1+ datasets, most rPPG-based methods can't achieve satisfactory results when training on these two datasets while testing on the CSMAD dataset. 

The experimental results on all three cross-dataset evaluations demonstrate that the proposed VMrPPG achieves the excellent generalization ability when facing unseen scenarios. It is also noted that the texture-based methods are suffered from huge performance degradation on cross-dataset evaluations, while the drop of the rPPG-based methods are insignificant, which again demonstrate the superior performance of rPPG-based methods on detecting 3D mask attacks.

\subsection{Ablation Studies}
To evaluate each component of the proposed method, a set of ablation studies are performed on the HKBU-Mars V2 dataset, which contains the greatest variations. 
The baseline method extracts the rPPG signals from the SeetaFace-aligned faces, encodes the signals on the spatial-temporal representation, and learns the features using a ResNet-18 as in \cite{niu2020rhythmnet}. The same evaluation protocol of 20 rounds of LOOCV \cite{liu2021multi} is applied here.

\subsubsection{Effects of Stitching-based Face Alignment}
\begin{figure}[ht]
	\centering
	\includegraphics[width=0.8\linewidth]{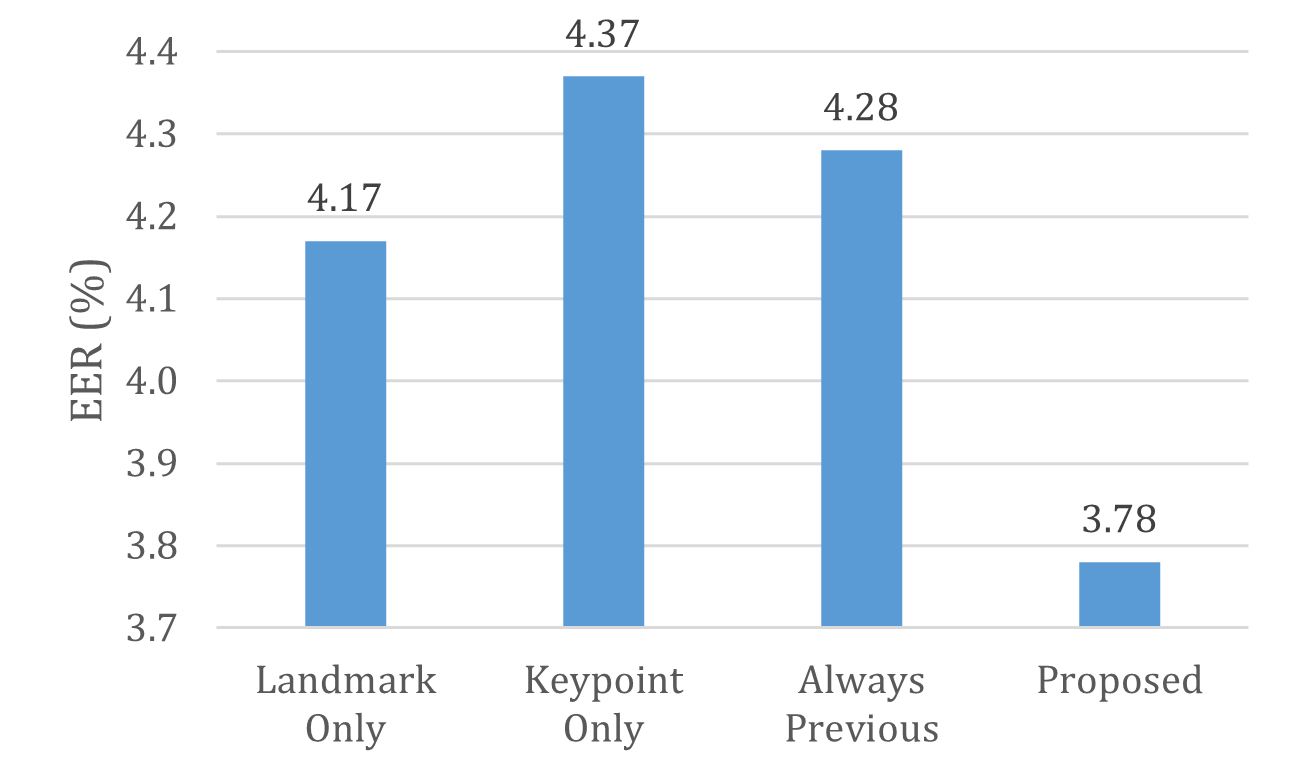}
	\caption{Ablation study on face alignment methods. The EER of the proposed stitching-based face alignment is $0.39\%$ lower than the baseline method.}
	\label{fig:ablation-alignment}
\end{figure}

The proposed stitching-based face alignment is compared with other face alignment methods. 
The proposed method utilizes both SIFT keypoints and facial landmarks for face alignment, and hence one compared method solely utilizes SIFT keypoints (\textbf{Keypoint Only}) and another solely utilizes facial landmarks (\textbf{Landmark Only}). As the proposed method utilizes dynamic programming to select intermediate faces to build a face alignment chain, another compared method fixes the intermediate face at the previous frame (\textbf{Always Previous}). Other components of the baseline method remain unchanged. The experimental results are shown in Fig. \ref{fig:ablation-alignment}.

With the same classifier and parameter settings, it is observed that the proposed stitching-based face alignment contributes the most distinct rPPG signals. Compared to the baseline fixing the intermediate face at the first frame and using only landmarks for alignment, the proposed method reduces the EER from $4.17\%$ to $3.78\%$. The method using only SIFT keypoints and the one fixing the intermediate face at the previous frame fall behind, which presents the negative effects caused by the error propagation over a long alignment chain when facing notable pose variations.

\subsubsection{Effects of rPPG-based Face Anti-spoofing Framework}
\label{sec:ablation-framework}
The improvements of the proposed VMrPPG are assessed progressively. The baseline feature representation is the spatial-temporal representation (\textbf{STR}) encoding rPPG signals from faces after stitching-based face alignment. The weighted multi-channel STR (\textbf{WMC-STR}) expands the features from the RGB color space to the RGB, YUV and Lab color spaces, and weights the features via a facial vascular density mask. The proposed lightweight EfficientNet with GRU (\textbf{ENetGRU}) is employed to replace the baseline ResNet-18 and extract features from \textbf{WMC-STR}. The \textbf{STR}, \textbf{WMC-STR}, and \textbf{ENetGRU} are applied progressively to evaluate the performance gain by including each of the proposed building blocks.

\begin{figure}[!t]
	\centering
	\includegraphics[width=0.8\linewidth]{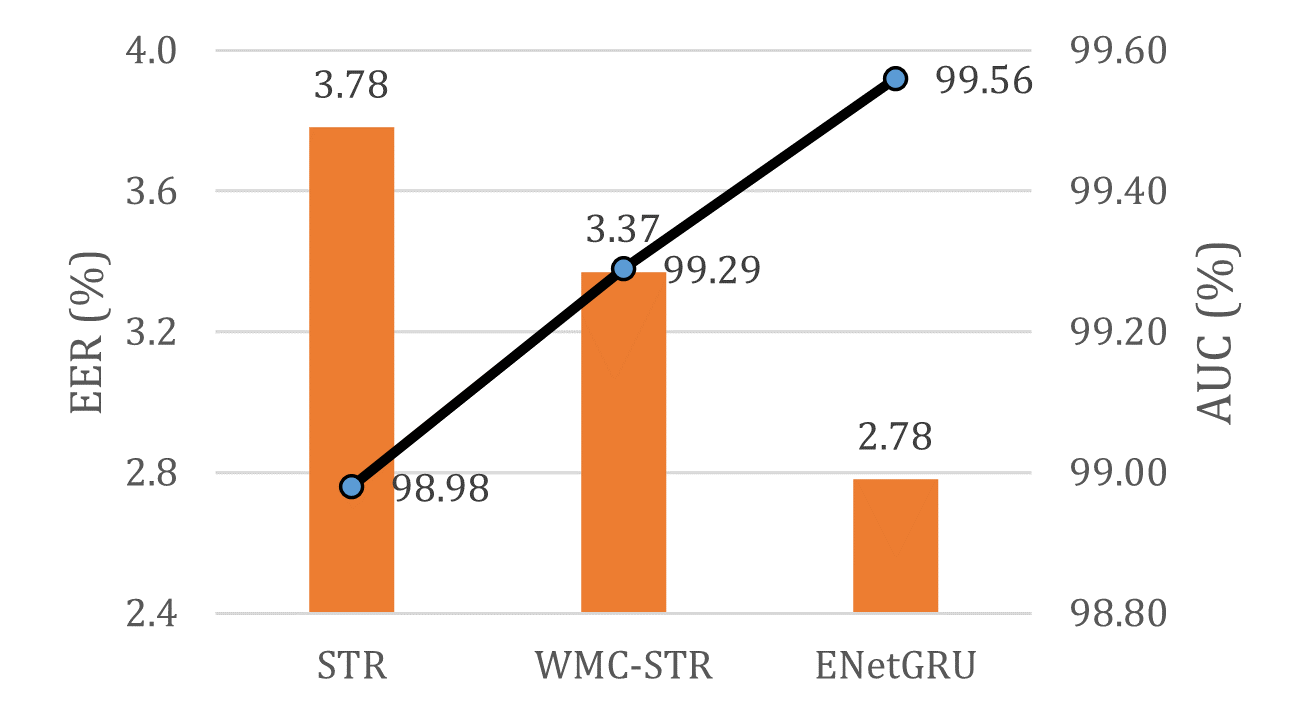}
	\caption{Ablation study on rPPG-based face anti-spoofing framework in terms of the EER and the AUC. The EER gradually decreases by $0.41\%$ and $0.59\%$, and the AUC progressively increases by $0.31\%$ and $0.27\%$ when using weighted multi-channel STR (\textbf{WMC-STR}) and replacing ResNet-18 with the proposed lightweight EfficientNet with GRU (\textbf{ENetGRU}).}
	\label{fig:ablation-framework}
\end{figure}

The comparison results are shown in Fig. \ref{fig:ablation-framework}. By utilizing the spatial-temporal representation and ResNet-18 only, the EER and AUC are $3.78\%$ and $98.98\%$, respectively. By using the proposed weighted multi-channel STR, the EER decreases to $3.37\%$ and the AUC increases to $99.29\%$, which indicates that the characteristics of rPPG signals in multiple color spaces are helpful to create distinct signal representations, and highlighting rPPG signals in regions with dense facial vessels is helpful for spoofing detection. After replacing ResNet-18 by ENetGRU, the EER further decreases to $2.78\%$ and the AUC further increases to $99.56\%$, which suggests the advanced power of the proposed lightweight EfficientNet with GRU in capturing the discriminant features for spoofing detection.

\subsubsection{Effects of Using Multiple Color Spaces}
To illustrate the benefits of utilizing multiple color spaces, the proposed method is compared to methods utilizing single color space, \ie, RGB, YUV, HSV, Lab, and a learning-based color space LC$_1$C$_2$ \cite{lu2018color}. All the other components remain unchanged as STR in Section \ref{sec:ablation-framework}. The results are summarized in Table \ref{table:ablation-color-space}. 
It is witnessed from Table \ref{table:ablation-color-space} that \textbf{WMC-STR} performs better than methods using any single color space, including the learning-based color space. The results show that the complementary information residing in different color spaces help construct a more robust rPPG representation.

\begin{table}[!t]
    \caption{Ablation studies of utilizing rPPG signals in different color spaces.}
	\begin{center}
		\begin{tabular}{|l|c|c|}
			\hline
			Color Space & EER & AUC \\
			\hline
			RGB (STR) & $3.78$ & $98.98$ \\
			Lab & $3.78$ & $99.10$ \\
			YUV & $4.17$ & $98.84$ \\
                HSV & $13.19$ & $91.39$ \\
			LC$_1$C$_2$ \cite{lu2018color} & $3.81$ & $98.83$ \\
                \hline			
                WMC-STR $\quad\quad$ & $3.37$ & $99.29$ \\
			\hline
		\end{tabular}
	\end{center}
	\label{table:ablation-color-space}
\end{table}

\subsubsection{Impact of Dataset Variances}
\begin{figure}[!ht]
	\subfigure[ ]{\includegraphics[width=0.95\linewidth]{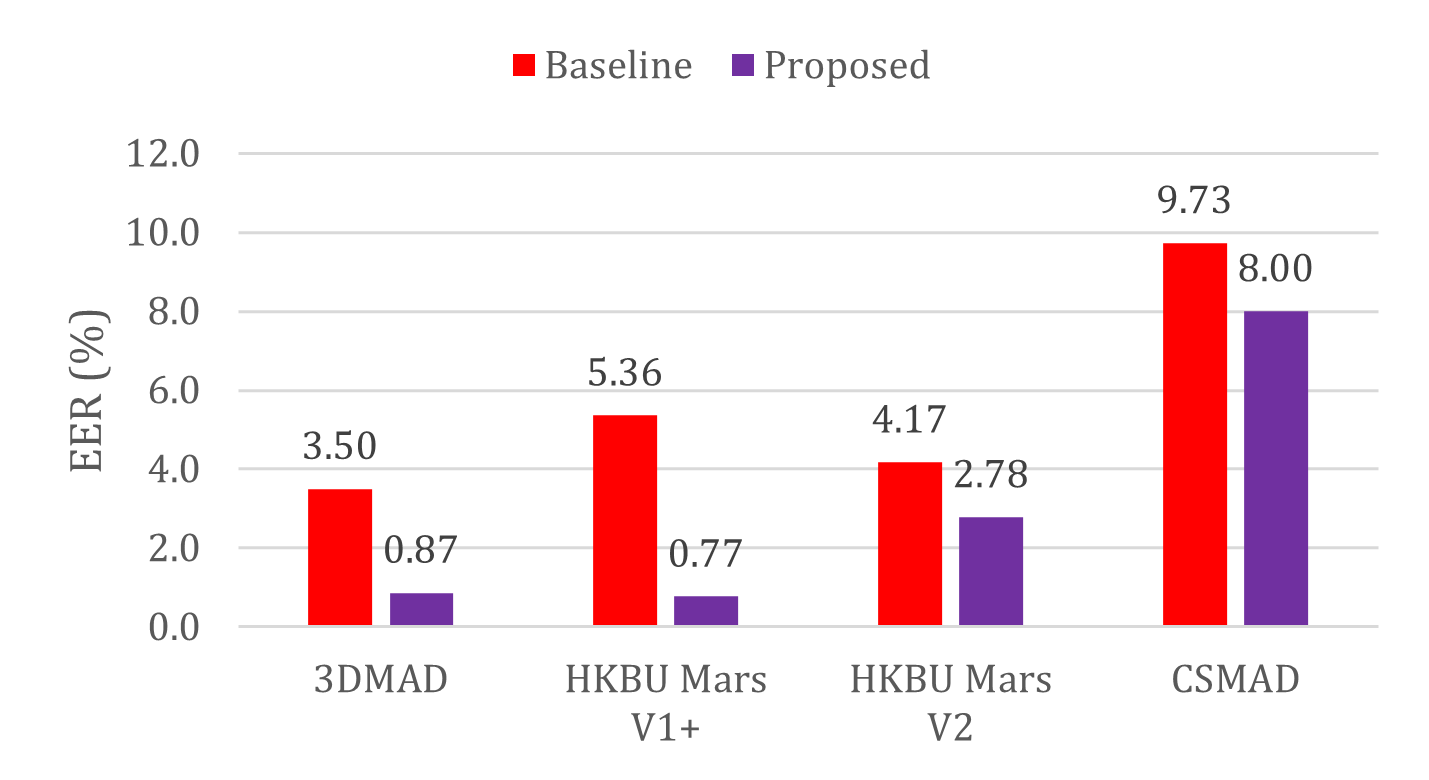}
		\label{fig:ablation-base-eer}}
	\centering
	\subfigure[ ]{\includegraphics[width=0.95\linewidth]{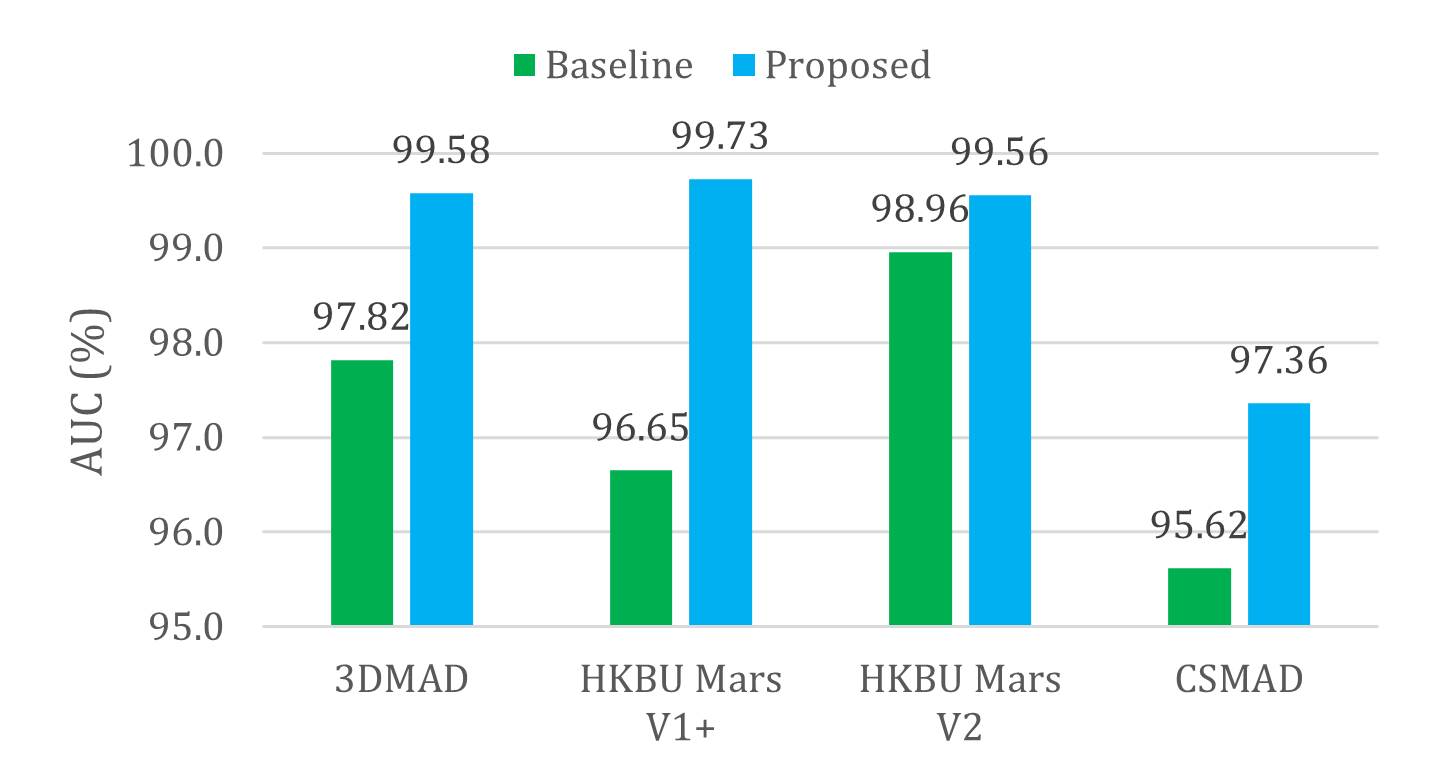}
		\label{fig:ablation-base-auc}}
	\caption{Ablation studies on dataset variances. The proposed VMrPPG consistently and significantly outperforms the baseline method on all four datasets in terms of both the EER and the AUC.}
	\label{fig:ablation-barchart}
\end{figure}

The proposed method is compared with the baseline method on four datasets, which does not have any proposed components,  and reserves landmark-only face alignment, spatial-temporal representation, and ResNet-18. The same evaluation protocol is used for each dataset as in intra-dataset evaluation. 
The comparisons in terms of the EER and the AUC are shown in Fig. \ref{fig:ablation-base-eer} and \ref{fig:ablation-base-auc}, respectively. Compared to the baseline method, the proposed method achieves a solid and consistent performance gain, \ie, $2.63\%$, $4.59\%$, $1.39\%$, and $1.73\%$ in terms of the EER on the 3DMAD, HKBU-Mars V1+, HKBU-Mars V2, and CSMAD datasets, respectively, and $1.76\%$, $3.08\%$, $0.60\%$, $1.74\%$ respectively in terms of the AUC.

\subsection{Experimental Results on Other Spoofing Attacks}
\label{sec:otherAttacks}
The rPPG signals can be used to detect various types of spoofing attacks from the recorded videos. To evaluate the ability of detecting other spoofing types, the proposed method is evaluated on the Idiap Replay Attack dataset \cite{chingovska2012effectiveness} for detecting photo attacks and video replay attacks, following the same experimental settings and the short-time observation (1 second) protocol as in LeTSrPPG \cite{liu2022learning}. 
The comparison results to the state-of-the-art rPPG-based and appearance-based methods are summarized in Table \ref{table:intra-replay-attack-1s}. 
It can be seen that the proposed VMrPPG significantly and consistently outperforms all the compared methods 
in detecting photo attacks and video replay attacks. Compared to the second-best method, MS-LBP \cite{erdogmus2014spoofing}, the proposed method uses less liveness cues but achieves better scores, with the reduction of 4.04\% on the HTER on the development set, 2.51\% on the HTER on the test set, 3.75\% on the EER, and an improvement of 1.31\% on the AUC. Compared with the previously best performed rPPG-based method, LeTSrPPG \cite{liu2022learning}, the proposed method achieves significant performance gains on all four evaluation metrics.
\begin{table}[!t]
    \caption{Comparison on the Idiap Replay Attack Dataset for short-time observations.}
	\begin{center}
	\resizebox{1\columnwidth}{!}{
		\begin{tabular}{|l|c|c|c|c|}
			\hline
			Method & HTER\_dev & HTER\_test & EER & AUC \\
			\hline
			$\blacktriangle$ MS-LBP \cite{erdogmus2014spoofing} & \underline{$8.54 \pm 1.40$} & \underline{$8.43 \pm 8.00$} & \underline{$8.76$} & \underline{$97.40$} \\
			$\star$ GrPPG \cite{li2016generalized} & $45.30 \pm 0.60$ & $45.30 \pm 5.20$ & $45.30$ & $56.50$ \\
			$\star$ PPGSecure \cite{nowara2017ppgsecure} & $39.10 \pm 0.60$ & $38.90 \pm 4.50$ & $39.10$ & $65.50$ \\
			$\star$ LrPPG \cite{liu20163d} & $44.20 \pm 0.60$ & $44.30 \pm 4.80$ & $44.20$ & $59.00$ \\
			$\star$ CFrPPG \cite{liu2018remote} & $36.30 \pm 0.70$ & $36.20 \pm 5.10$ & $36.30$ & $68.00$ \\
			$\star$ LeTSrPPG \cite{liu2022learning} & $18.50 \pm 0.70$ & $18.70 \pm 6.10$ & $18.60$ & $88.90$ \\
			\hline
			$\star$ Proposed VMrPPG & $\bm{4.50 \pm 0.40}$ & $\bm{5.92 \pm 4.13}$ & $\bm{5.01}$ & $\bm{98.71}$ \\
			\hline
		\end{tabular}}	
    \end{center}
    \label{table:intra-replay-attack-1s}
\end{table}

\subsection{Discussions}
The proposed VMrPPG aims to address the two challenges of existing rPPG-based methods: the alignment error of face sequences that may greatly distort the rPPG signals, and the weak and noisy rPPG signals. 
The ablation studies in Fig. \ref{fig:ablation-alignment} show that the proposed face alignment algorithm using both SIFT keypoints and facial landmarks mitigates the distortion in the rPPG signals caused by face alignment errors. The ablation studies in Fig. \ref{fig:ablation-framework}, Fig. \ref{fig:ablation-barchart}, and Table \ref{table:ablation-color-space} demonstrate that the proposed signal weighting mechanism based on the vascular density and the color space fusion helps construct a robust rPPG signal representation. As evidenced in Fig.~\ref{fig:ablation-framework}, the proposed customized EfficientNet with the GRU has strong discriminant power. To show the generalization ability of the proposed framework, we have conducted a series of experiments, \eg, the intra-dataset evaluations on the 3DMAD, HKBU-Mars V1+, HKBU-Mars V2 and CSMAD datasets, the intra-dataset evaluations for short-time observations on the 3DMAD, HKBU-Mars V1+ and Idiap Replay Attack datasets, and the cross-dataset evaluations among the 3DMAD, HKBU-Mars V1+, and CSMAD datasets. The proposed method also significantly outperforms all the compared methods for other spoofing attacks such as printed photo attacks and video-replay attacks, as demonstrated in Section \ref{sec:otherAttacks}. In summary, the proposed method significantly outperforms the state-of-the-art rPPG-based methods on multiple datasets under different evaluation protocols for different spoofing attacks. 

\section{Conclusion}
rPPG signals provide an effective liveness clue to detect 3D mask attacks. However, the rPPG signal has low signal to noise ratio and is sensitive to the spatial positions of video frames. Thus, the facial micro movements and inaccurate face alignment, though can be tolerated by a face recognition system, largely weaken the rPPG signal and hence make it ineffective. To address this challenge, we propose a landmark-anchored face stitching algorithm to align the face at a pixel-wise level, design a vascular-weighted multi-channel spatial-temporal representation to rPPG signals, and extract reliable spatial-temporal features by a lightweight EfficientNet with a GRU. More precisely, our contributions are four-fold: 1) To align the face at the pixel level to enhance the rPPG signal quality, a landmark-anchored face stitching algorithm is proposed 
which utilizes the facial landmarks as anchor points to prevent the error propagation in the face alignment chain, and utilizes face stitching through keypoint to achieve an accurate and consistent face alignment. 
2) The rPPG signal features are extracted from different color spaces to make use of the signal characteristics embedded in different color spaces. 3) The processed signals from each ROI are stacked as a spatial-temporal representation, and then weighted using the density of the facial blood vessels, to highlight the ROIs with rich blood vessels. 4) 
The lightweight EfficientNet with the GRU following the compound-scaling mechanism is developed for spatial-temporal feature learning. The proposed method is compared with state-of-the-art rPPG-based face anti-spoofing models under both intra-dataset and cross-dataset evaluations on five datasets. 
Experimental results show that the proposed approach significantly and consistently outperforms all the compared rPPG-based methods.





{\footnotesize
\balance
\bibliographystyle{IEEEtran}
\bibliography{paper}

\begin{thebibliography}{10}
\providecommand{\url}[1]{#1}
\csname url@samestyle\endcsname
\providecommand{\newblock}{\relax}
\providecommand{\bibinfo}[2]{#2}
\providecommand{\BIBentrySTDinterwordspacing}{\spaceskip=0pt\relax}
\providecommand{\BIBentryALTinterwordstretchfactor}{4}
\providecommand{\BIBentryALTinterwordspacing}{\spaceskip=\fontdimen2\font plus
\BIBentryALTinterwordstretchfactor\fontdimen3\font minus
  \fontdimen4\font\relax}
\providecommand{\BIBforeignlanguage}[2]{{%
\expandafter\ifx\csname l@#1\endcsname\relax
\typeout{** WARNING: IEEEtran.bst: No hyphenation pattern has been}%
\typeout{** loaded for the language `#1'. Using the pattern for}%
\typeout{** the default language instead.}%
\else
\language=\csname l@#1\endcsname
\fi
#2}}
\providecommand{\BIBdecl}{\relax}
\BIBdecl

\bibitem{boulkenafet2015face}
Z.~Boulkenafet, J.~Komulainen, and A.~Hadid, ``Face anti-spoofing based on
  color texture analysis,'' in \emph{IEEE International Conference on Image
  Processing (ICIP)}.\hskip 1em plus 0.5em minus 0.4em\relax IEEE, 2015, pp.
  2636--2640.

\bibitem{chen2020attention}
H.~{Chen}, G.~{Hu}, Z.~{Lei}, Y.~{Chen}, N.~M. {Robertson}, and S.~Z. {Li},
  ``Attention-based two-stream convolutional networks for face spoofing
  detection,'' \emph{IEEE Transactions on Information Forensics and Security},
  vol.~15, pp. 578--593, 2020.

\bibitem{shen2019facebagnet}
T.~Shen, Y.~Huang, and Z.~Tong, ``{FaceBagNet}: Bag-of-local-features model for
  multi-modal face anti-spoofing,'' in \emph{IEEE Conference on Computer Vision
  and Pattern Recognition Workshops (CVPRW)}, 2019, pp. 1611--1616.

\bibitem{sun2020face}
W.~{Sun}, Y.~{Song}, C.~{Chen}, J.~{Huang}, and A.~C. {Kot}, ``Face spoofing
  detection based on local ternary label supervision in fully convolutional
  networks,'' \emph{IEEE Transactions on Information Forensics and Security},
  vol.~15, pp. 3181--3196, 2020.

\bibitem{wang2021unsupervised}
G.~{Wang}, H.~{Han}, S.~{Shan}, and X.~{Chen}, ``Unsupervised adversarial
  domain adaptation for cross-domain face presentation attack detection,''
  \emph{IEEE Transactions on Information Forensics and Security}, vol.~16, pp.
  56--69, 2021.

\bibitem{erdogmus2014spoofing}
N.~{Erdogmus} and S.~{Marcel}, ``Spoofing face recognition with {3D} masks,''
  \emph{IEEE Transactions on Information Forensics and Security}, vol.~9,
  no.~7, pp. 1084--1097, 2014.

\bibitem{kose2013shape}
N.~{Kose} and J.~{Dugelay}, ``Shape and texture based countermeasure to protect
  face recognition systems against mask attacks,'' in \emph{IEEE Conference on
  Computer Vision and Pattern Recognition Workshops (CVPRW)}, 2013, pp.
  111--116.

\bibitem{steiner2016reliable}
H.~{Steiner}, A.~{Kolb}, and N.~{Jung}, ``Reliable face anti-spoofing using
  multispectral {SWIR} imaging,'' in \emph{International Conference on
  Biometrics (ICB)}, 2016, pp. 1--8.

\bibitem{shao2017deep}
R.~{Shao}, X.~{Lan}, and P.~C. {Yuen}, ``Deep convolutional dynamic texture
  learning with adaptive channel-discriminability for {3D} mask face
  anti-spoofing,'' in \emph{IEEE International Joint Conference on Biometrics
  (IJCB)}, 2017, pp. 748--755.

\bibitem{shao2018joint}
R.~Shao, X.~Lan, and P.~C. Yuen, ``Joint discriminative learning of deep
  dynamic textures for {3D} mask face anti-spoofing,'' \emph{IEEE Transactions
  on Information Forensics and Security}, vol.~14, no.~4, pp. 923--938, 2018.

\bibitem{siddiqui2016face}
T.~A. {Siddiqui}, S.~{Bharadwaj}, T.~I. {Dhamecha}, A.~{Agarwal}, M.~{Vatsa},
  R.~{Singh}, and N.~{Ratha}, ``Face anti-spoofing with multifeature videolet
  aggregation,'' in \emph{International Conference on Pattern Recognition
  (ICPR)}, 2016, pp. 1035--1040.

\bibitem{tang2016shape}
Y.~Tang and L.~Chen, ``Shape analysis based anti-spoofing {3D} face recognition
  with mask attacks,'' in \emph{International Workshop on Representations,
  Analysis and Recognition of Shape and Motion From Imaging Data}.\hskip 1em
  plus 0.5em minus 0.4em\relax Springer, 2016, pp. 41--55.

\bibitem{wang2018face}
Y.~Wang, S.~Chen, W.~Li, D.~Huang, and Y.~Wang, ``Face anti-spoofing to {3D}
  masks by combining texture and geometry features,'' in \emph{Chinese
  Conference on Biometric Recognition}.\hskip 1em plus 0.5em minus 0.4em\relax
  Springer, 2018, pp. 399--408.

\bibitem{birla2022patron}
L.~{Birla} and P.~{Gupta}, ``{PATRON}: Exploring respiratory signal derived
  from non-contact face videos for face anti-spoofing,'' \emph{Expert Systems
  with Applications}, vol. 187, no. 115883, 2022.

\bibitem{birla2022sunrise}
L.~{Birla}, P.~{Gupta}, and S.~{Kumar}, ``{SUNRISE}: Improving {3D} mask face
  anti-spoofing for short videos using pre-emptive split and merge,''
  \emph{IEEE Transactions on Dependable and Secure Computing}, vol.~20, no.~3,
  pp. 1927--1940, 2023.

\bibitem{li2016generalized}
X.~{Li}, J.~{Komulainen}, G.~{Zhao}, P.~C. {Yuen}, and M.~{Pietik{\"a}inen},
  ``Generalized face anti-spoofing by detecting pulse from face videos,'' in
  \emph{International Conference on Pattern Recognition (ICPR)}.\hskip 1em plus
  0.5em minus 0.4em\relax IEEE, 2016, pp. 4244--4249.

\bibitem{liu20163d}
S.~{Liu}, P.~C. {Yuen}, S.~{Zhang}, and G.~{Zhao}, ``{3D} mask face
  anti-spoofing with remote photoplethysmography,'' in \emph{European
  Conference on Computer Vision (ECCV)}.\hskip 1em plus 0.5em minus 0.4em\relax
  Springer, 2016, pp. 85--100.

\bibitem{liu2018remote}
S.~{Liu}, X.~{Lan}, and P.~C. {Yuen}, ``Remote photoplethysmography
  correspondence feature for {3D} mask face presentation attack detection,'' in
  \emph{European Conference on Computer Vision (ECCV)}, 2018, pp. 558--573.

\bibitem{liu2020temporal}
------, ``Temporal similarity analysis of remote photoplethysmography for fast
  {3D} mask face presentation attack detection,'' in \emph{IEEE Winter
  Conference on Applications of Computer Vision (WACV)}, 2020, pp. 2597--2605.

\bibitem{liu2021multi}
------, ``Multi-channel remote photoplethysmography correspondence feature for
  {3D} mask face presentation attack detection,'' \emph{IEEE Transactions on
  Information Forensics and Security}, vol.~16, pp. 2683--2696, 2021.

\bibitem{liu2022learning}
------, ``Learning temporal similarity of remote photoplethysmography for fast
  {3D} mask face presentation attack detection,'' \emph{IEEE Transactions on
  Information Forensics and Security}, vol.~17, pp. 3195--3210, 2022.

\bibitem{nowara2017ppgsecure}
E.~M. {Nowara}, A.~{Sabharwal}, and A.~{Veeraraghavan}, ``{PPGSecure}:
  Biometric presentation attack detection using photopletysmograms,'' in
  \emph{IEEE International Conference on Automatic Face and Gesture Recognition
  (FG)}, 2017, pp. 56--62.

\bibitem{yao2021rppgbased}
C.~{Yao}, S.~{Wang}, J.~{Zhang}, W.~{He}, H.~{Du}, J.~{Ren}, R.~{Bai}, and
  J.~{Liu}, ``{rPPG}-based spoofing detection for face mask attack using
  efficientnet on weighted spatial-temporal representation,'' in \emph{IEEE
  International Conference on Image Processing (ICIP)}, 2021.

\bibitem{yu2021transrppg}
Z.~{Yu}, X.~{Li}, P.~{Wang}, and G.~{Zhao}, ``{TransRPPG}: Remote
  photoplethysmography transformer for {3D} mask face presentation attack
  detection,'' \emph{IEEE Signal Processing Letters (SPL)}, vol.~28, pp.
  1290--1294, 2021.

\bibitem{george2020biometric}
A.~{George}, Z.~{Mostaani}, D.~{Geissenbuhler}, O.~{Nikisins}, A.~{Anjos}, and
  S.~{Marcel}, ``Biometric face presentation attack detection with
  multi-channel convolutional neural network,'' \emph{IEEE Transactions on
  Information Forensics and Security}, vol.~15, pp. 42--55, 2020.

\bibitem{liu2019deep}
Y.~{Liu}, J.~{Stehouwer}, A.~{Jourabloo}, and X.~{Liu}, ``Deep tree learning
  for zero-shot face anti-spoofing,'' in \emph{IEEE Conference on Computer
  Vision and Pattern Recognition (CVPR)}, 2019, pp. 4680--4689.

\bibitem{liu2020disentangling}
Y.~{Liu}, J.~{Stehouwer}, and X.~{Liu}, ``On disentangling spoof trace for
  generic face anti-spoofing,'' in \emph{European Conference on Computer Vision
  (ECCV)}.\hskip 1em plus 0.5em minus 0.4em\relax Springer, 2020, pp. 406--422.

\bibitem{yu2020nasfas}
Z.~{Yu}, J.~{Wan}, Y.~{Qin}, X.~{Li}, S.~Z. {Li}, and G.~{Zhao}, ``{NAS-FAS}:
  Static-dynamic central difference network search for face anti-spoofing,''
  \emph{IEEE Transactions on Pattern Analysis and Machine Intelligence},
  vol.~43, no.~9, pp. 3005--3023, 2021.

\bibitem{liu2022contrastive}
A.~{Liu}, C.~{Zhao}, Z.~{Yu}, J.~{Wan}, A.~{Su} \emph{et~al.}, ``Contrastive
  context-aware learning for {3D} high-fidelity mask face presentation attack
  detection,'' \emph{IEEE Transactions on Information Forensics and Security},
  vol.~17, pp. 2497--2507, 2022.

\bibitem{bhattacharjee2017what}
S.~{Bhattacharjee} and S.~{Marcel}, ``What you can't see can help
  you—extended-range imaging for {3D}-mask presentation attack detection,''
  in \emph{International Conference of the Biometrics Special Interest Group
  (BIOSIG)}, 2017, pp. 1--7.

\bibitem{alsufyani2018biometric}
N.~{Alsufyani}, A.~{Ali}, S.~{Hoque}, and F.~{Deravi}, ``Biometric presentation
  attack detection using gaze alignment,'' in \emph{IEEE International
  Conference on Identity, Security, and Behavior Analysis (ISBA)}, 2018, pp.
  1--8.

\bibitem{niu2019robust}
X.~{Niu}, X.~{Zhao}, H.~{Han}, A.~{Das}, A.~{Dantcheva}, S.~{Shan}, and
  X.~{Chen}, ``Robust remote heart rate estimation from face utilizing
  spatial-temporal attention,'' in \emph{International Conference on Automatic
  Face Gesture Recognition (FG)}, 2019, pp. 1--8.

\bibitem{niu2020rhythmnet}
X.~{Niu}, S.~{Shan}, H.~{Han}, and X.~{Chen}, ``Rhythmnet: end-to-end heart
  rate estimation from face via spatial-temporal epresentation,'' \emph{IEEE
  Transactions on Image Processing}, vol.~29, pp. 2409--2423, 2020.

\bibitem{lowe2004distinctive}
D.~G. Lowe, ``Distinctive image features from scale-invariant keypoints,''
  \emph{International Journal of Computer Vision (IJCV)}, vol.~60, no.~2, pp.
  91--110, 2004.

\bibitem{wang2017algorithmic}
W.~{Wang}, A.~C. {den Brinker}, S.~{Stuijk}, and G.~{de Haan}, ``Algorithmic
  principles of remote {PPG},'' \emph{IEEE Transactions on Biomedical
  Engineering}, vol.~64, no.~7, pp. 1479--1491, 2017.

\bibitem{sun2018colour}
G.~{Sun}, H.~{Li}, and B.~{Li}, \emph{Colour Atlas of Human Blood Vessels Cast
  and Angiography}.\hskip 1em plus 0.5em minus 0.4em\relax Shenyang, Liaoning:
  Liaoning Science and Technology Publishing House, 2018.

\bibitem{jia2020survey}
S.~Jia, G.~Guo, and Z.~Xu, ``A survey on {3D} mask presentation attack
  detection and countermeasures,'' \emph{Pattern Recognition}, vol.~98, no.
  107032, 2020.

\bibitem{erdogmus20133dmad}
N.~{Erdogmus} and S.~{Marcel}, ``Spoofing in {2D} face recognition with {3D}
  masks and anti-spoofing with {Kinect},'' \emph{Biometrics: Theory,
  Applications and Systems (BTAS)}, pp. 1--6, 2013.

\bibitem{kannala2012bsif}
J.~Kannala and E.~Rahtu, ``{BSIF}: Binarized statistical image features,'' in
  \emph{IEEE International conference on pattern recognition (ICPR)}, 2012, pp.
  1363--1366.

\bibitem{liu2016hkbu-mars-v2}
S.~{Liu}, B.~{Yang}, P.~C. {Yuen}, and G.~{Zhao}, ``A {3D} mask face
  anti-spoofing database with real world variations,'' in \emph{IEEE Conference
  on Computer Vision and Pattern Recognition Workshops (CVPRW)}, 2016, pp.
  100--106.

\bibitem{liu2018learning}
Y.~{Liu}, A.~{Jourabloo}, and X.~{Liu}, ``Learning deep models for face
  anti-spoofing: Binary or auxiliary supervision,'' in \emph{IEEE Conference on
  Computer Vision and Pattern Recognition (CVPR)}, 2018, pp. 389--398.

\bibitem{hamdan2018self}
B.~Hamdan and K.~Mokhtar, ``A self-immune to {3D} masks attacks face
  recognition system,'' \emph{Signal, Image and Video Processing}, vol.~12,
  no.~6, pp. 1053--1060, 2018.

\bibitem{li2014remote}
X.~{Li}, J.~{Chen}, G.~{Zhao}, and M.~{Pietik{\"a}inen}, ``Remote heart rate
  measurement from face videos under realistic situations,'' in \emph{IEEE
  Conference on Computer Vision and Pattern Recognition (CVPR)}, 2014, pp.
  4264--4271.

\bibitem{heusch2018pulse}
G.~Heusch and S.~Marcel, ``Pulse-based features for face presentation attack
  detection,'' in \emph{IEEE International Conference on Biometrics Theory,
  Applications and Systems (BTAS)}, 2018, pp. 1--8.

\bibitem{lin2019face}
B.~Lin, X.~Li, Z.~Yu, and G.~Zhao, ``Face liveness detection by {rPPG} features
  and contextual patch-based {CNN},'' in \emph{International Conference on
  Biometric Engineering and Applications}, 2019, pp. 61--68.

\bibitem{yu2023deep}
Z.~{Yu}, Y.~{Qin}, X.~{Li}, C.~{Zhao}, Z.~{Lei}, and G.~{Zhao}, ``Deep learning
  for face anti-spoofing: A survey,'' \emph{IEEE Transactions on Pattern
  Analysis and Machine Intelligence}, vol.~45, no.~5, pp. 5609--5631, 2023.

\bibitem{liu2023spoof}
Y.~{Liu} and X.~{Liu}, ``Spoof trace disentanglement for generic face
  anti-spoofing,'' \emph{IEEE Transactions on Pattern Analysis and Machine
  Intelligence}, vol.~45, no.~3, pp. 3813--3830, 2023.

\bibitem{liu2021face}
A.~{Liu}, Z.~{Tan}, J.~{Wan}, Y.~{Liang}, Z.~{Lei}, G.~{Guo}, and S.~Z. {Li},
  ``Face anti-spoofing via adversarial cross-modality translation,'' \emph{IEEE
  Transactions on Information Forensics and Security}, vol.~16, pp. 2759--2772,
  2021.

\bibitem{qin2022meta}
Y.~{Qin}, Z.~{Yu}, L.~{Yan}, Z.~{Wang}, C.~{Zhao}, and Z.~{Lei}, ``Meta-teacher
  for face anti-spoofing,'' \emph{IEEE Transactions on Pattern Analysis and
  Machine Intelligence}, vol.~44, no.~10, pp. 6311--6326, 2022.

\bibitem{agarwal2017face}
A.~Agarwal, D.~Yadav, N.~Kohli, R.~Singh, M.~Vatsa, and A.~Noore, ``Face
  presentation attack with latex masks in multispectral videos,'' in \emph{IEEE
  Conference on Computer Vision and Pattern Recognition Workshops (CVPRW)},
  2017, pp. 81--89.

\bibitem{haan2013robust}
G.~{de Haan} and V.~{Jeanne}, ``Robust pulse rate from chrominance-based
  {rPPG},'' \emph{IEEE Transactions on Biomedical Engineering}, vol.~60,
  no.~10, pp. 2878--2886, 2013.

\bibitem{wang2016novel}
W.~{Wang}, S.~{Stuijk}, and G.~{de Haan}, ``A novel algorithm for remote
  photoplethysmography: Spatial subspace rotation,'' \emph{IEEE Transactions on
  Biomedical Engineering}, vol.~63, no.~9, pp. 1974--1984, 2016.

\bibitem{lovisotto2020seeing}
G.~{Lovisotto}, H.~{Turner}, S.~{Eberz}, and I.~{Martinovic}, ``Seeing red:
  {PPG} biometrics using smartphone cameras,'' in \emph{IEEE Conference on
  Computer Vision and Pattern Recognition Workshops (CVPRW)}, 2020, pp.
  3565--3574.

\bibitem{wu2017funnel}
S.~Wu, M.~Kan, Z.~He, S.~Shan, and X.~Chen, ``Funnel-structured cascade for
  multi-view face detection with alignment-awareness,'' \emph{Neurocomputing},
  vol. 221, pp. 138--145, 2017.

\bibitem{geng2011face}
C.~{Geng} and X.~{Jiang}, ``Face recognition based on the multi-scale local
  image structures,'' \emph{Pattern Recognition}, vol.~44, no. 10-11, pp.
  2565--2575, 2011.

\bibitem{tan2019efficientnet}
M.~{Tan} and Q.~{Le}, ``{EfficientNet}: Rethinking model scaling for
  convolutional neural networks,'' in \emph{International Conference on Machine
  Learning (ICML)}, 2019, pp. 6105--6114.

\bibitem{comas2021turnip}
A.~{Comas}, T.~K. {Marks}, H.~{Mansour}, S.~{Lohit}, Y.~{Ma}, and X.~{Liu},
  ``Turnip: Time-series {U-Net} with recurrence for {NIR} imaging {PPG},'' in
  \emph{IEEE International Conference on Image Processing (ICIP)}, 2021, pp.
  309--313.

\bibitem{sandler2018mobilenetv2}
M.~{Sandler}, A.~{Howard}, M.~{Zhu}, A.~{Zhmoginov}, and L.~C. {Chen},
  ``{MobileNetv2}: Inverted residuals and linear bottlenecks,'' in \emph{IEEE
  Conference on Computer Vision and Pattern Recognition (CVPR)}, 2018, pp.
  4510--4520.

\bibitem{hu2018squeeze}
J.~{Hu}, L.~{Shen}, and G.~{Sun}, ``Squeeze-and-excitation networks,'' in
  \emph{IEEE Conference on Computer Vision and Pattern Recognition (CVPR)},
  2018, pp. 7132--7141.

\bibitem{bhattacharjee2018spoofing}
S.~Bhattacharjee, A.~Mohammadi, and S.~Marcel, ``Spoofing deep face recognition
  with custom silicone masks,'' in \emph{IEEE 9th International Conference on
  Biometrics Theory, Applications and Systems (BTAS)}.\hskip 1em plus 0.5em
  minus 0.4em\relax IEEE, 2018, pp. 1--7.

\bibitem{chingovska2012effectiveness}
I.~{Chingovska}, A.~{Anjos}, and S.~{Marcel}, ``On the effectiveness of local
  binary patterns in face anti-spoofing,'' in \emph{IEEE International
  Conference of Biometrics Special Interest Group (BIOSIG)}, 2012, pp. 1--7.

\bibitem{lu2018color}
Z.~{Lu}, X.~{Jiang}, and A.~{Kot}, ``Color space construction by optimizing
  luminance and chrominance components for face recognition,'' \emph{Pattern
  Recognition}, vol.~83, pp. 456--468, 2018.

\end{thebibliography}
}
\end{document}